\documentclass[10pt,twocolumn,letterpaper]{article}

\usepackage{wacv}              

\usepackage{soul}
\sethlcolor{yellow} 

\definecolor{bubblegum}{rgb}{0.99, 0.76, 0.8}
\definecolor{darkpink}{rgb}{0.91, 0.33, 0.5}
\definecolor{pastelred}{rgb}{1.0, 0.41, 0.38}
\definecolor{scarlet}{rgb}{1.0, 0.13, 0.0}

\newcommand{\modelname}{GAEA}
\newcommand{\datasetname}{GAEA-1.4M}
\newcommand{\dataseteval}{GAEA-Bench}
\newcommand{\dec}[1]{$_{\textcolor{scarlet}{-#1}}$}
\definecolor{wacvblue}{rgb}{0.21,0.49,0.74}
\definecolor{codegreen}{rgb}{0,0.6,0}
\usepackage[pagebackref,breaklinks,colorlinks,allcolors=wacvblue]{hyperref}

\usepackage{multirow}
\usepackage{arydshln}
\usepackage{tcolorbox}
\usepackage{colortbl}

\usepackage{multirow}
\usepackage{graphicx}
\usepackage{pifont}
\newcommand{\smartparagraph}[1]{\vspace{1.5pt} \noindent {\bf #1}}
\usepackage{mdframed} 
\newcommand{\myNum}[1]{(\emph{#1})}

\def\wacvPaperID{762} 
\def\confName{WACV}
\def\confYear{2026}

\title{GAEA: A Geolocation Aware Conversational Assistant}

\author{
    \begin{minipage}[t]{\textwidth}
        \centering
        {Ron Campos}\textsuperscript{ \large{\thanks{equally contributing first authors}}},
        {Ashmal Vayani}\textsuperscript{ *}, 
        {Parth Parag Kulkarni}\textsuperscript{ *},
        {Rohit Gupta}, \\
        {Aizan Zafar},
        {Aritra Dutta},
        {Mubarak Shah} \\ [0.5em]
        \large{
            Center for Research in Computer Vision, University of Central Florida
        }
    \end{minipage}
}

\begin{document}
\maketitle
\begin{abstract}
Image geolocalization, in which an AI model traditionally predicts the precise GPS coordinates of an image, is a challenging task with many downstream applications. However, the user cannot utilize the model to further their knowledge beyond the GPS coordinates; the model lacks an understanding of the location and the conversational ability to communicate with the user. In recent days, with the tremendous progress of large multimodal models (LMMs)---proprietary and open-source---researchers have attempted to geolocalize images via LMMs. However, the issues remain unaddressed; beyond general tasks, for more specialized downstream tasks, such as geolocalization, LMMs struggle. In this work, we propose solving this problem by introducing a conversational model, \textnormal{\modelname}, that provides information regarding the location of an image as the user requires. No large-scale dataset enabling the training of such a model exists. Thus, we propose \textnormal{\datasetname}, a comprehensive dataset comprising over 800k images and approximately 1.4M question-answer pairs, constructed by leveraging OpenStreetMap (OSM) attributes and geographical context clues. For quantitative evaluation, we propose a diverse benchmark, \textnormal{GAEA-Bench}, comprising 3.5k image-text pairs to evaluate conversational capabilities equipped with diverse question types. We consider 11 state-of-the-art open-source and proprietary LMMs and demonstrate that \textnormal{\modelname} significantly outperforms the best open-source model, LLaVA-OneVision, by 18.2\% and the best proprietary model, GPT-4o, by 7.2\%. 
Our dataset, model and codes are available \href{https://ucf-crcv.github.io/GAEA/}{\textbf{\textcolor{purple}{https://ucf-crcv.github.io/GAEA}.}}
\end{abstract}

\begin{figure}[t]
    \centering
    \includegraphics[width=\linewidth]{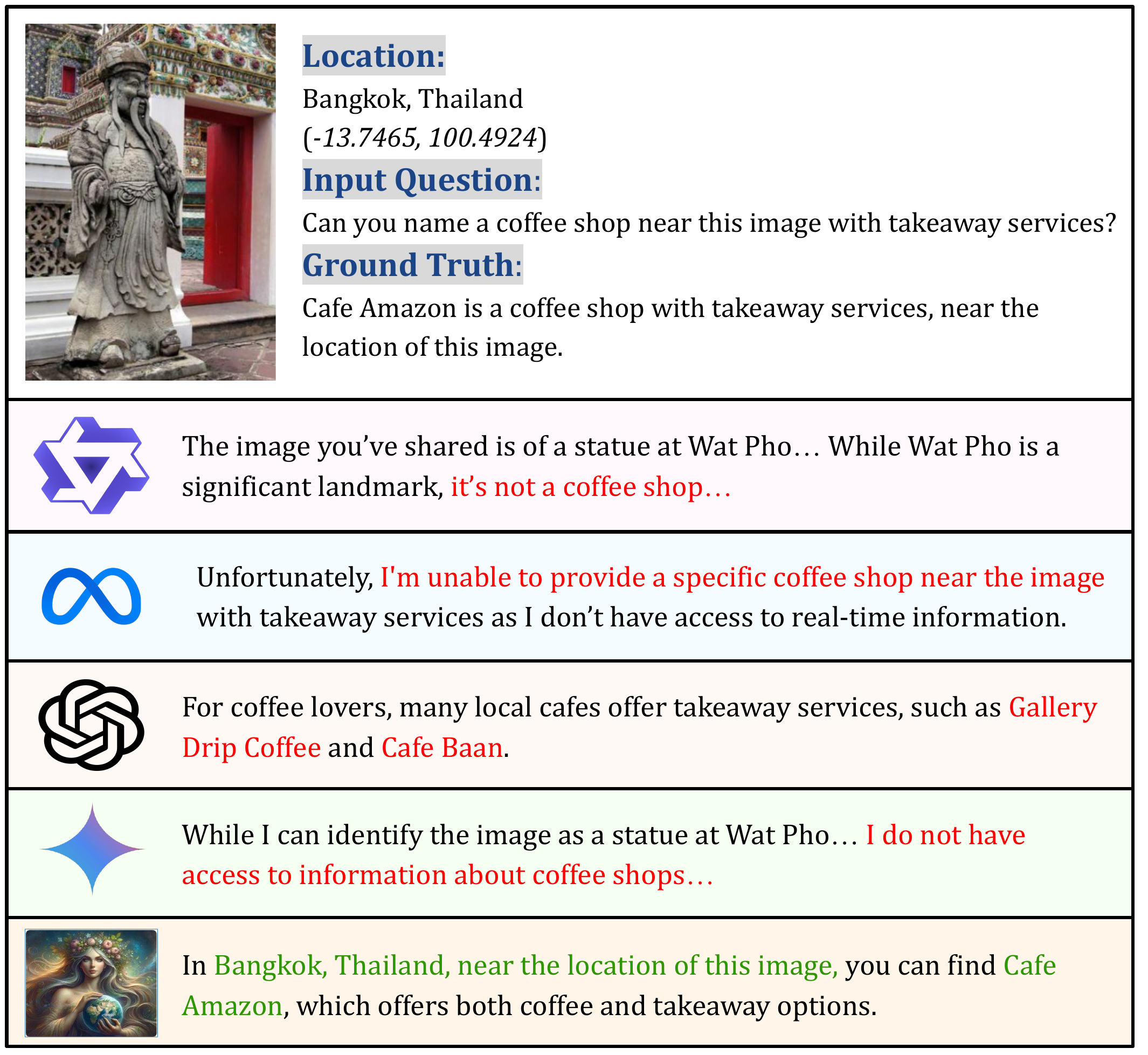}

    \caption{We compare the performance of various LMMs on the geographically‑grounded visual‑question‑answering task, included in our new GAEA-Bench benchmark. Most LMMs can describe the Wat Pho statue, but only GAEA, our Geolocation AwarE Assistant, retrieves the correct nearby cafe, \textit{Cafe Amazon}.}
    \vspace{-2em}
    \label{fig:TEASER-GAEA}
\end{figure}
\vspace{-1em}
\section{Introduction}
\label{sec:intro}

Image geolocalization \cite{transgeo,vivanco2024geoclip,haas2024pigeon,georeasoner-icml25,yan2024georeasoner} is a notoriously challenging task, in which, conventional AI models predict the precise GPS coordinate of an image taken anywhere on Earth.~Moreover, seasonal changes, geographical and climatic diversity, changes in solar zenith angle, and lack of diverse image distributions make the geolocalization task more challenging. Although challenging, geolocalization has direct applications in multiple domains, including tourism \cite{chalvatzaras2022survey}, navigation, urban planning \cite{shen2017streetvizor}, and security \cite{vivanco2024geoclip, dutta2024multiview}, among others.

Recently, the CLIP-inspired image-to-GPS retrieval approach, GeoCLIP \cite{vivanco2024geoclip}, has shown significant performance in global-scale image geolocalization. To further mitigate the performance gap, and to increase the generalization capacity of the models, interestingly, a new wave of works infuse {\em human-level cognition and inference capacity} in their model training \cite{haas2024pigeon, georeasoner-icml25,yan2024georeasoner}.~E.g., PIGEON is trained on data from the popular geolocalization game GeoGuesser \cite{geoguessr}; a recent vision-language model, GeoReasoner \cite{georeasoner-icml25} uses user- and administrator-maintained approximately 3K textual clues from GeoGuessr and Tuxun gaming platforms.

These focused geolocalization models lack a geographical understanding of the predicted locations beyond their GPS coordinates. They cannot provide additional information that might be invaluable for applications such as tourism, navigation, urban planning, etc. Even if the models possess that understanding, they do not possess the conversational ability to convey that information and fail to meet the user's needs. In contrast, despite having the conversational capability, visually and textually prompted large language models (LLMs) \cite{llama3, qwen2.5, thawakar2024mobillama} and their multimodal variants, popularly referred to as large multimodal models (LMMs) \cite{gpt3, llama32vision, team2023gemini, flamingo, maaz2024palo}, fail to capture fine-grained nuances from an image in specialized downstream tasks such as geolocalization, making their predictions vastly imprecise and worse than random guesses in many cases; see \Cref{fig:TEASER-GAEA}. 
\begin{figure*}[t]
  \centering
  \includegraphics[width=\textwidth]{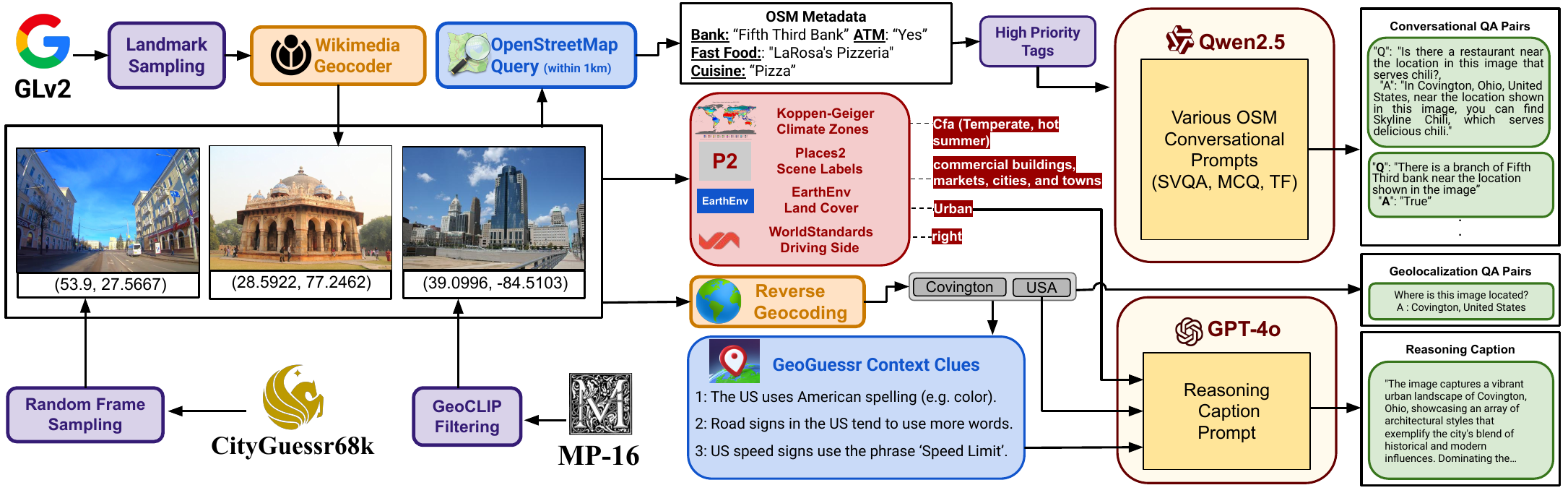}
    \vspace{-1.2em}

\caption{\small{\textbf{Data Collection and Annotation Pipeline.} \datasetname\ includes geographically diverse visual samples from various data sources, such as MP-16 \cite{larson2017benchmarking}, GLD-v2 \cite{weyand2020google}, and CityGuesser68k \cite{kulkarni2024cityguessr} \textit{(left)}. We also incorporate additional metadata and auxiliary context for each image from OpenStreetMap (OSM), ranging from climate zones to geographical clues about the country \textit{(middle)}. Using open-source LLMs and GPT-4o, we generate four diverse question-answer pairs across geolocation, reasoning, and conversational subsets \textit{(right)}.}}
  \label{fig:GeoLLM_Flow}
\end{figure*}

Motivated by these aspects, in this paper, we propose \textnormal{\modelname}, an open-source conversational model with global-scale geolocalization capability. To the best of our knowledge, this is the first work in the ground-view geolocalization domain that introduces an open-source conversational chatbot, where the user can obtain image geolocalization, relevant description of the image, and engage in a meaningful conversation about the surrounding landmarks, natural attractions, restaurants or coffee shops, medical or emergency facilities, and recreational areas. 

However, training an open-source LMM with conversational capacity is not straightforward. These models are data-hungry, and their training is computationally intensive. Unfortunately, no dataset can facilitate the training of such a model. To this end, we meticulously curate a \datasetname\ ---a high-quality conversational VQA pair equipped with diversity in scene understanding and image captions for training and instruction tuning the LMMs on the street-level geolocalization task. \datasetname\ is a comprehensive dataset consisting over 800k images from \textit{MP-16} \cite{larson2017benchmarking}, \textit{GLD-v2} \cite{weyand2020google}, and CityGuessr68k \cite{kulkarni2024cityguessr} covering locations around the Earth. We augment these images with rich metadata from the OpenStreetMap (OSM) \cite{osm} at a 1 km radius, a first effort of its kind. OSM attributes contain details about the surrounding area, nearby landmarks, accessible services, and historical buildup of the region. Additionally, the geolocalizable explanatory captions set contains 385K image-QA pairs and is equipped with their corresponding knowledge and reasoning captions. These knowledge and reasoning captions are constructed using a set of geographical context clues from GeoGuessr \cite{geoguessr} that enable the model to gain a holistic understanding of the location. Taken together, \textnormal{\datasetname} is the largest and most comprehensive collection of geolocalizable and conversational QA pairs. Using this, we train our conversational chatbot, \textnormal{\modelname}.

To quantitatively evaluate the conversational capability of LMMs and address the scarcity of benchmark datasets in a geolocalization setting, we propose \textnormal{\dataseteval}, a diverse set of 3.5k conversational question samples. 

\modelname-bench comprises multiple-choice (MCQs) and true/false (T/Fs) for checking a model's understanding and choosing capability, short questions (SVQAs) for testing a model's knowledge, and long questions (LVQAs) for evaluating a model's descriptive and in-depth explanation ability about the location in question.

We summarize the main contributions as follows:
\begin{itemize}
    \item We propose \datasetname\ (\S\Cref{sec:dataset}), a new dataset for training conversational image geolocalization models.
    \item For evaluating conversational capabilities in a geolocalization setting (\S\Cref{sec:experiment}), we propose GAEA-Bench, a novel benchmark of 3.5k samples with various question types, including short, long, MCQs, and T/F. 
    \item We propose GAEA, a conversational chatbot (\S\ref{sec:method}) that extends beyond global-scale geolocalization, providing information about the location described by an image.
    \item We quantitatively compare the performance of our model to 8 state-of-the-art open-source and 3 proprietary LMMs, including GPT-4o \cite{gpt4} and Gemini-2.0-Flash \cite{team2023gemini}.
\end{itemize}

\section{Related Work}\label{sec:related}

\smartparagraph{Large Multimodal Models}\label{sec:related_LMM}
(LMMs) have been at the forefront of computer vision research; geo-localizable LMMs are in their nascent stages.
Multimodal learning unifies different modalities by a common representation. 
By {\em contrastively} fitting text and images into the same feature space, CLIP~\cite{clip} has revolutionized multimodal learning. LLMs like GPT2~\cite{gpt2} made strides in representing text and next token prediction. Visual question answering~(VQA) was of interest before, but after LLaVA~\cite{llava} and BLIP2~\cite{blip2} combined the conversational aspects of LLMs and the representational capabilities of CLIP-like models, many problems of VQA are addressed. 
After that, numerous modern works emerged, such as GeoChat~\cite{geochat}, Qwen2.5-VL~\cite{Qwen2.5-VL}, LLaMA-3.2 Vision~\cite{llama32vision}, and LLaVA-OneVision~\cite{llavaov} as well as proprietary models like GPT4~\cite{gpt4} and Gemini~\cite{team2023gemini}. Although most of these models are excellent for general VQA, they perform poorly on specialized tasks in fields like medicine, statistics, and geolocalization. This inspires the need for specialized LMMs that can address specific tasks. 

\smartparagraph{Geo-localization}\label{sec:related_geo} is a crucial field in vision research with essential applications in forensics, social media, and exploration; see \cite{chalvatzaras2022survey, shen2017streetvizor, vivanco2024geoclip}. On a global scale, Weyand et al. \cite{weyand2016planet} first introduced a classification-based approach on the Im2GPS \cite{hays2008im2gps} dataset. Vo et al.\ \cite{vo2017revisiting} introduced classification in multiple hierarchies, while CPlaNet \cite{cplanet} introduced a combinatorial partitioning technique for combining coarse hierarchies to predict finer ones. Over the years many other works like ISNs \cite{isns}, TransGeo \cite{transgeo}, TransLocator \cite{translocator}, and GeoDecoder \cite{geocoder} have made significant advancements in this classification-based worldwide geolocalization by introducing scene-based specialized encoders and hierarchical evaluation, auxiliary scene recognition, and twin encoder approach, and a query-based encoder-decoder architecture, respectively. PIGEON \cite{haas2024pigeon}, the most recent work, leverages the image representation capabilities of the CLIP vision encoder and a unique clustering approach to improve its geo-localization performance. Image-to-image retrieval models tend to be more accurate than their classification-based counterparts, but are infeasible on a global scale due to their requirement for large reference image galleries. GeoCLIP \cite{vivanco2024geoclip} was the first work to incorporate the contrastive multimodal learning between images and raw GPS information that revolutionized this domain by introducing a more accurate retrieval-based model for a global scale.

These specialized models work well for worldwide image geo-localization but lack the conversational aspect that can aid an individual in gaining a holistic understanding of a location portrayed in an image. GeoReasoner \cite{georeasoner-icml25} attempts to incorporate an inherent geospatial understanding into a LMM by looking at specific information displayed in the image. It also introduces the idea of {\em locatability}, which can determine the extent of that information present in the image, which may improve the reasoning capability of the model. The model, however, lacks the {\em conversational aspect}, and the locatability-based filtering of data might hurt its generalization capability. We address these issues in \textnormal{\modelname} by primarily focusing on its conversational ability.

The generalizability of \textnormal{\modelname} comes from its training data. All specialized geo-localization models that function on a global scale train their model on MP-16 \cite{larson2017benchmarking}, which is a large-scale worldwide dataset. However, it lacks the verbal context required in LMM training. Hence, we introduce a new conversational dataset \textnormal{\datasetname}; see details in \S\Cref{sec:dataset}. Additionally, we introduce the first conversational benchmark in \S\Cref{sec:method} to evaluate Geolocalization LMMs and an evaluation pipeline to judge the efficacy of such models.
\begin{figure*}[t]
  \centering
  \includegraphics[width=\textwidth]{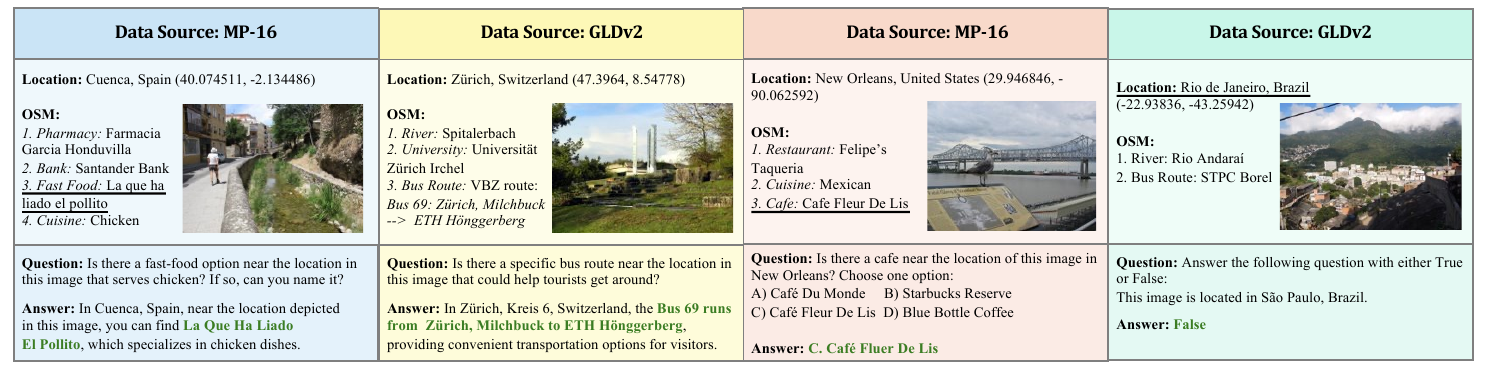}
    \vspace{-1.2em}
\caption{\textbf{Qualitative samples} from our \datasetname\ training set showcasing various question-types, including multiple-choice, true/false, short and long VQAs generated using a proprietary model, GPT-4o \cite{OpenAI2024}. We carefully select geographical tags from OSM metadata to generate conversational question-answer pairs.}
  \label{fig:qual_osm}
  \vspace{-3mm}
\end{figure*}

\section{Dataset Curation and Annotation}\label{sec:dataset}
\subsection{GAEA-1.4M}\label{sec:gaea1.4m}

The \datasetname\ dataset provides comprehensive global coverage, featuring both rich conversational and diverse geolocalization sets. It includes various QA formats, such as MCQs, True/False, and open-ended VQA (long and short), from more than 234 countries/territories, grouped under conversational and geolocalization groups. Spanning 40k cities across 7 continents, \textnormal{\datasetname} is structured into two key groups: conversational and geolocalization. With over 1.4 million QA pairs, it captures the geographical diversity of both underrepresented and widely recognized regions worldwide. \Cref{fig:GeoLLM_Flow} shows our complete curation pipeline, which we will discuss piecewise.


\smartparagraph{Acquiring Diverse Geo-localizable Images.}~We sample geographically  diverse visual data from MediaEval 2016 (MP-16) \cite{larson2017benchmarking}, Google Landmarks v2 (GLDv2) \cite{weyand2020google}, and CityGuessr \cite{kulkarni2024cityguessr} to curate \datasetname.

\smartparagraph{MP-16} contains over 4.6 million geotagged Flickr images, including indoor and outdoor scenes. For our street-view geolocalization subset, we filter out indoor images, retaining 3 million outdoor images. However, some of these images are non-geolocalizable, such as close-up shots of doors, grass, or wires, which are excluded from the final dataset. To filter out non-geolocalizable images, we process all 3 million samples using GeoCLIP \cite{vivanco2024geoclip}, which is trained on the full MP-16 dataset and effectively identifies non-geolocalizable outlier images. GeoCLIP assigns a confidence score based on its ability to predict GPS coordinates, with higher scores indicating geo-localizability. We set a confidence threshold of 0.75 and computed the distance between the ground truth MP-16 GPS coordinates and the GeoCLIP’s predicted location. We retain the images if this distance is less than 500 km; see additional ablations on different thresholds and distance metrics in the Appendix.

To achieve a balanced geographical distribution in \textnormal{\datasetname}, we use the 10th hierarchy of S2-Cells \cite{s2cells} to partition our filtered MP-16 dataset into 16,753 spatial grid cells. S2-Cells enable hierarchical spatial indexing, ensuring diverse global coverage while preventing overrepresenting densely imaged regions. We randomly sample up to 200 images from each cell, resulting in a final set of over 750k distinct samples.

\smartparagraph{GLDv2} \cite{weyand2020google} is a fine-grained landmark recognition dataset featuring natural and human-made landmarks across diverse time zones, climates, and lighting conditions. Given the significance of landmark geolocation for real-world applications, we randomly sample 50K distinct landmarks from GLDv2. These highly recognizable landmarks offer rich geographic and cultural context. Each image is linked to Wikipedia metadata, from which we extract GPS coordinates using the \texttt{Wikimedia} \cite{wikimedia} API. We then apply the \texttt{reverse\_geocoder} Python library to determine each landmark's corresponding city and country.

\smartparagraph{CityGuessr68k} \cite{kulkarni2024cityguessr} focuses on global video-based geolocalization emphasizing urban regions and hierarchical prediction across 166 major cities. To incorporate this diversity, we randomly sample one frame from each of the 54k training videos and include them in our dataset. These three sources provide over 852k geographically diverse geolocalizable images, forming \textnormal{\datasetname} dataset.

\subsubsection{Meta-data curation for dataset annotation}
After acquiring all visual samples for our GAEA-Conversational Assistant, we churn the metadata for each image for a comprehensive QA-pair generation. 

\smartparagraph{Churning OSM metadata.} OpenStreetMap (OSM) \cite{osm} is a collaborative open-source mapping platform that provides extensive geographical data. In our work, OSM plays a central role by enriching geolocalization and conversational capabilities. We retrieve metadata from a 1 km radius around the GPS coordinates of 850k images, leveraging OSM’s detailed, publicly annotated tags. These tags cover many real-world elements, including amenities, transportation, hotels, and restaurants, making them invaluable for our ground-view geolocalization and QA generation. 

OSM data is multilingual, which is a key challenge. To ensure accessibility, we use GPT-4o \cite{gpt4} to translate these annotations into English. Additionally, many retrieved tags consisted of plain numbers or non-meaningful entries, which we systematically filtered out to retain only informative and contextually relevant metadata. To our knowledge, this is the first work to utilize OSM’s rich metadata to develop a conversational chatbot for ground-view geolocalization. Figure \ref{fig:qual_osm} shows high-priority meta-tags derived from OSM churned data.

\smartparagraph{Curating Country-Specific Geographical Clues.} We web-crawled diverse clues from \texttt{Plonkit}\cite{plonkit}, an open-source community resource for the GeoGuessr \cite{geoguessr} game, which has over 65 million players. Similar datasets have been used in recent works \cite{haas2024pigeon, georeasoner-icml25}. We obtained 129 country clues but found gaps for some countries, such as New Zealand and France. To address this, we curated clues for 58 additional countries using GPT-4o, aligning them with \texttt{Plonkit}'s style, resulting in altogether 187 countries. These clues are incorporated into our dataset for generating reasoning-based QAs. For examples of the type of clues utilized, see \Cref{fig:clues} in the Supplementary material.

\smartparagraph{Additional Metadata.} For auxiliary context, we group our country-specific, geographically diverse dataset in 31 Köppen-Geiger climate zones \cite{beck2018present}.~We obtain the traffic direction data through WorldStandards \cite{ws} and \textit{Land Cover Use} statistics from EarthEnv \cite{earthenv}. Additionally, we compute scene labels for each image using the Places2 \cite{zhou2017places} database.

\begin{figure*}[t]
  \centering
  \includegraphics[width=\textwidth]{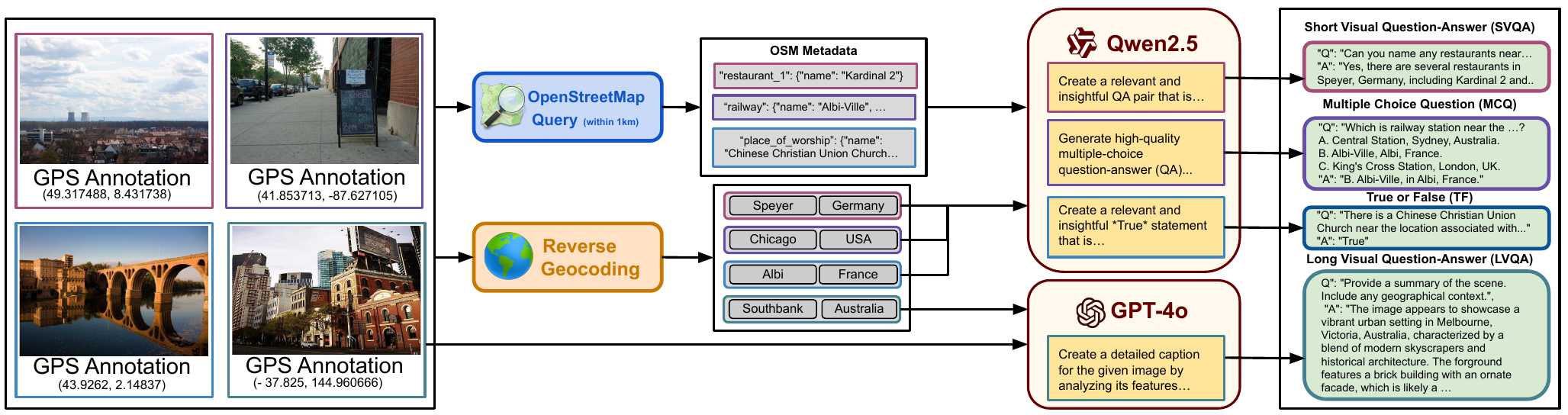}

\caption{\small{\textbf{Overview of GAEA-Bench.} GAEA-Bench is designed to evaluate the conversational abilities of various LMMs across different question types, including MCQs, T/F, and both short and long VQAs. We have carefully selected a subset of 3.5k samples from MP-16 \cite{larson2017benchmarking} and generated corresponding OSM metadata to generate QA pairs using GPT-4o \cite{OpenAI2024}. GAEA-Bench aims to fill the gap in conversational benchmarks by incorporating geolocalization capabilities.}}
  \label{fig:GeoLLM-Bench}
  \vspace{-4mm}
\end{figure*}
\subsubsection{Question-Answer (QA) Pairs Generation}
\datasetname\ is carefully curated to enhance ground-view geolocalization through diverse, context-rich QA pairs; see \Cref{fig:GeoLLM_Flow}. Comprising over 800k distinct images and around 1.4 million QA pairs, it stands as the largest and most comprehensive dataset for this task; see \Cref{fig:qual_osm}. Unlike existing works, such as \cite{georeasoner-icml25, dou2024gaga}, which are limited to \texttt{JSON} structures and fewer question types, our work emphasizes the conversational capabilities of the model, providing a broader range of QA formats. The dataset is divided into three subsets---Conversational, Reasoning, and Geo-Localization, each designed to capture different aspects of geographic understanding. These subsets feature various question formats, including multiple-choice \cite{narnaware2025sb}, true/false, and open-ended questions (SVQA and LVQA) \cite{vayani2024all}. Below, we detail the curation process for each subset.

\smartparagraph{Conversational QA Generation.} We generate conversational QA pairs using OSM metadata from the sampled MP-16 and GLDv2 subsets. We prompt Qwen-2.5-14B \cite{qwen2.5} with enriched OSM attributes to create diverse question formats, including short-form, multiple-choice, and true/false questions. These OSM tags cover various categories such as amenities, food places, financial institutions, government offices, accommodation, transportation, healthcare, religious sites, education, and waterways. This subset comprises over 380k questions. Figure \ref{fig:qual_osm} showcases qualitative samples of QA pairs curated using OSM churned data.

\smartparagraph{Geolocalization Questions.} To enhance the geolocalization capabilities of our GAEA model, we introduce large-scale meta-geographic information through geolocation-specific QA pairs. This subset consists of 820k image-question pairs designed to help the model predict the correct location of an image. We curate 50k geolocation questions from GLDv2, each corresponding to a distinct landmark, leveraging their global recognition to improve location-based reasoning. Additionally, we incorporate 54K geolocation QA pairs from CityGuessr, which focuses on urban environments, and 720k from MP-16, ensuring broad geographic coverage. This results in a diverse and well-distributed geolocation QA dataset spanning 234 countries and territories, 40k cities, and 7 continents.

\smartparagraph{Reasoning Questions.} We generate detailed image-caption QA pairs (Long-VQA) to enhance fine-grained reasoning in our GAEA model. We prompt GPT-4o \cite{gpt4} with each image, its scene labels, and country-specific geographical attributes, including GeoGuessr clues, traffic-side driving information, Köppen-Geiger climate zone, and land cover data. While scene labels are unique to each image, the other attributes provide country-level context. GPT-4o integrates this information to generate contextually rich and highly correlated captions with the provided geographic labels. These reasoning-based captions strengthen the model’s geolocalization and conversational capabilities and induce a rich semantic understanding in our model by infusing {\em human-level cognition and inference capability}, enabling the model to emphasize why particular image features might be associated with specific geographical contexts, reducing disinformation \cite{raza2025vldbench, raza2025responsible}. In total, we curate 237k knowledge-driven LVQA pairs.

\vspace{-1mm}


\subsection{GAEA-Bench}
Existing benchmarks for evaluating geolocalization tasks mainly focus on retrieval and classification-based methods, such as IM2GPS \cite{hays2008im2gps}, IM2GPS3k \cite{vo2017revisiting}, and GSW15k \cite{clark2023we}, which assess the distance between ground-truth and predicted GPS coordinates. However, there is a lack of conversational benchmarking datasets to evaluate the geolocalization and conversational capabilities of LMMs. We introduce \textnormal{\dataseteval}, a geographically diverse and conversationally rich multimodal benchmark to address these shortcomings. \textnormal{\dataseteval} is designed to assess LMMs across various question types, including MCQs, true/false, and long and short VQAs while integrating geolocalization tasks. It includes 3.5k image-text QA pairs that provide a rich geographical context for each image.
\begin{figure}[t]
  \centering
  \includegraphics[width=0.5\textwidth]{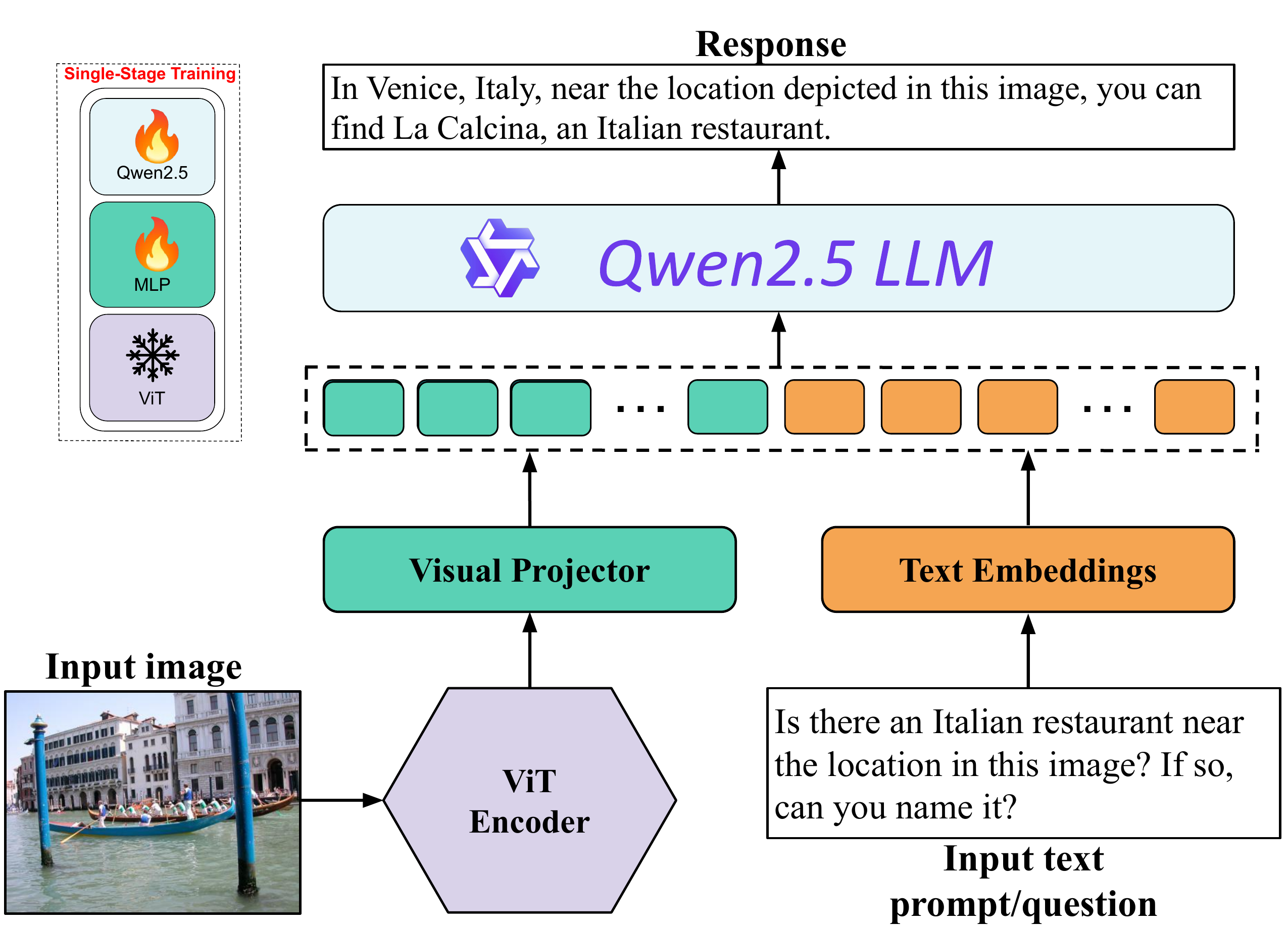}
    \vspace{-2em}
  \caption{\textbf{Overview of the \modelname\ model architecture and workflow.} An input image is first processed by a Vision Transformer (ViT) encoder, whose output is projected through a visual projector to obtain visual embeddings. Simultaneously, the input text prompt is converted into text embeddings. The combined visual and textual embeddings are then fed into the Qwen2.5 LLM space, which generates a response based on the multimodal input. We follow the single-stage training approach, unfreezing MLP, and performing LoRA fine-tuning in the same stage.}
  \label{fig:chatbot_training}
  \vspace{-5mm}
\end{figure}

\smartparagraph{GAEA-Bench Curation.} We curate a non-overlapping subset of highly geolocalizable MP-16 images, manually filtering out the non-geolocalizable ones. Using OpenStreetMaps (OSM), we generate metadata within a 1km radius and curate 975 short-form (SVQA), 978 multiple-choice (MCQ), and 978 true/false (T/F) questions. For long-form questions (LVQAs), we follow a similar process for generating reasoning questions in \textnormal{\dataseteval}, resulting in an additional 383 questions. In total, we curate 3,314 diverse image-text QA pairs. To ensure that the \textnormal{\dataseteval} remains independent of the training set, we select geographically distinct locations for its 3.5k samples. We show our GAEA-Bench annotation and curation process in \Cref{fig:GeoLLM-Bench}. The OSM metadata are fetched for each image and are passed to Qwen2.5-14B for generating several QA pairs, including SVQA, MCQ, and T/F.

\section{Model Architecture}\label{sec:method}
GAEA follows the architecture of the open-source model, Qwen2.5-VL \cite{Qwen2.5-VL}, which seamlessly integrates (1) a vision encoder, (2) a vision-to-language projector, and (3) a language model. The re-engineered vision-transformer (ViT) architecture incorporates 2D-RoPE and window attention. The projector is a two-layer multi-layer perception (MLP) to align raw patch features from the ViT and provides the final representation $\mathbf{E}^{\rm Joint}$ by concatenating the image embeddings,
$\mathbf{E}^{\rm Img}$ with the text embeddings, $\mathbf{E}^{\rm Text}$ such that $\mathbf{E}^{\rm Joint} = [\mathbf{E}^{\rm Img},  \mathbf{E}^{\rm Text}]$; see \Cref{fig:chatbot_training}.

\smartparagraph{Training Details.}~We perform single-stage fine-tuning of Qwen2.5VL on our GAEA Conversational Assistant dataset. The model is trained across all three subsets—\textit{geolocalization, reasoning, and conversational}—covering both open-ended QA formats (short and long answers) and decision-based questions (multiple-choice and true/false). This fine-tuning process enables the model to integrate rich geographical cues, contextual metadata, and image-specific attributes, enhancing its spatial reasoning, location inference, and multimodal conversational capabilities. We employ LoRA fine-tuning \cite{lora} with a rank of $r=16$ and $\alpha=32$ along with the unfrozen vision-to-language MLP projector. To handle varying image resolutions, we apply dynamic resolution processing: Images below $448\times448$ are upsampled, while those exceeding $1000\times 1000$ are downsampled, similar to \cite{Qwen2.5-VL}. The model is trained for one epoch over 12,600 steps.

\vspace{-1mm}
\section{Benchmarking and Evaluations}\label{sec:experiment}


\textnormal{\datasetname} training set comprises four distinct question types: Multiple Choice Questions (MCQs), True/False (T/F), and Short and Long Visual Question Answering (VQA). GAEA is meticulously trained to ensure conversational fluency while possessing the capability to geolocalize visual samples. Current evaluation frameworks primarily focus on standard geo-localization datasets, measuring accuracy using distance-based metrics at various scales, including Street (1 km), City (25 km), Region (200 km), Country (750 km), and Continent (2,500 km). However, these methods fail to assess the conversational capabilities of LMMs. To address this gap, we define our evaluation process in three key dimensions: (a) Conversational accuracy, (b) Quantitative geo-localization accuracy, and (c) Classification accuracy.

\subsection{Evaluation and Metrics}\label{sec:eval}
\textbf{Conversational Evaluation.}\label{sec:eval_conv}
Most geolocation-specific models operate as ``black box" systems, providing GPS coordinates without offering any reasoning or justification behind their outputs. In contrast, GAEA is the first model of its kind, explicitly trained on 1.4 million instructions, which include a significant number of knowledge-reasoning question-answer pairs. This enables GAEA to integrate world knowledge, such as geographical clues, conversational meta-tags, and advanced reasoning capabilities, making its geolocation predictions more transparent and insightful. To address the challenges of complex conversational evaluation, we benchmark 12 state-of-the-art open-source and closed-source LMMs on GAEA-Bench, meticulously curated to evaluate LMMs on diverse question types, including multiple-choice, true/false, and open-ended questions (short and long VQAs). See \S\Cref{sec:baselines} for the baselines used in this work. 

\begin{figure}[t]
  \centering
  \includegraphics[width = \columnwidth]{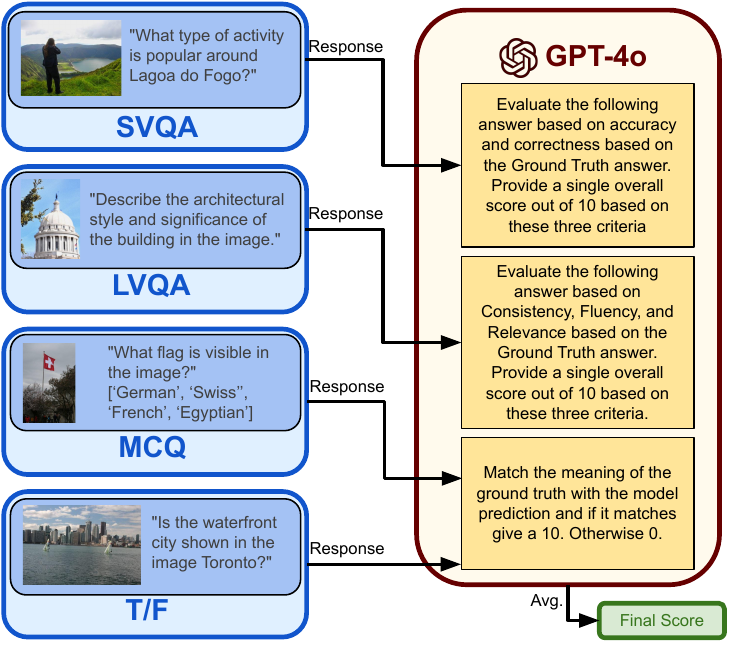}
   \caption{\textbf{Evaluation pipeline} for conversational benchmarking on GAEA-Bench, highlighting various question types we introduce in our \textnormal{GAEA-Bench}. Each question type is evaluated with various defined criteria using GPT-4o as a judge. For instance, SVQA is evaluated against \textit{accuracy} and \textit{correctness}, and LVQA is evaluated on \textit{Consistency}, \textit{Fluency}, and \textit{Relevancy} criteria.}
\label{fig:conv_pipe}   
\vspace{-4mm}
\end{figure}

We employ different prompts for each type of question. We use GPT-4o as a judge and prompt it to score responses to various types of questions with different criteria. We use \textit{accuracy} for MCQs and T/F, \textit{correctness} for SVQA, and \textit{consistency, relevance, and geographical correctness} for long VQAs (LVQAs); see the evaluation pipeline in \Cref{fig:conv_pipe}. Here, \textit{correctness} refers to how closely the model's output matches the location and the correct answer in the ground-truth response \cite{vayani2024all}. For LVQA, the \textit{consistency} metric evaluates the fluency and readability of the model's prediction \cite{sai2022survey, thawakar2024mobillama, vayani2024all}, while \textit{geographical correctness} assesses whether the model's prediction accurately identifies the correct city and country, directly matching the ground-truth answer. This is further discussed in \S\Cref{sec:appendix_prompt}, and \Cref{fig:prompts}.

\vspace{-1mm}

\smartparagraph{Quantitative Geo-localization Evaluation.}\label{sec:eval_quant}
We compared the performance of GAEA against six state-of-the-art (SoTA) geo-localization models, namely PlaNet \cite{weyand2016planet}, CPlaNet \cite{cplanet}, ISNs \cite{isns}, TransLocator \cite{translocator}, GeoDecoder \cite{clark2023we}, and PIGEON \cite{haas2024pigeon} on three standard geo-localization benchmarks including IM2GPS \cite{hays2008im2gps}, IM2GPS3k \cite{vo2017revisiting}, GWS15k \cite{clark2023we}. We prompt various LMMs to output the corresponding city and country to which the image belongs. We retrieve GPS coordinates using GeoPy \cite{geopy} and compute distance with ground truth. We compare the output with distance thresholds of 1 km, 25 km, 200 km, 750 km, and 2,500 km; see \Cref{tab:benchmark_results}.

\begin{figure}
  \centering
  \includegraphics[width = \columnwidth]{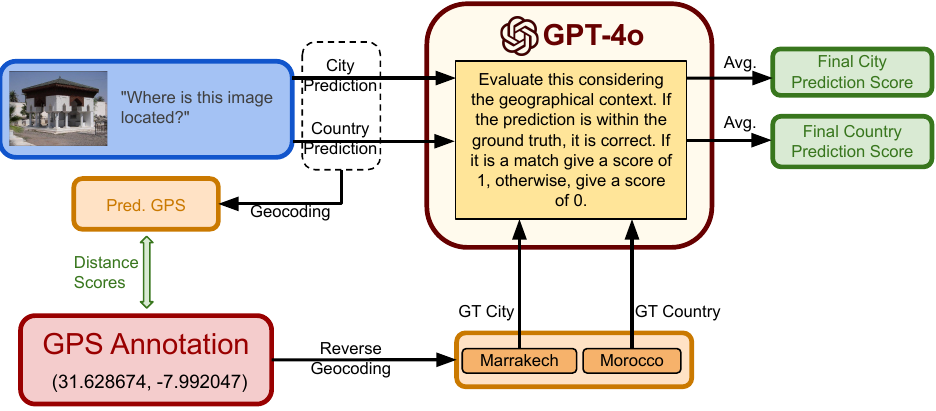}
   \caption{\textbf{Classification and distance threshold accuracy computation pipeline} simultaneously evaluates geolocalization performance at \textit{city} and \textit{country} level by comparing model predictions with ground truth annotations derived from reverse-geocoding GPS coordinates and accuracy at different distance thresholds by geocoding predictions of the model.}
\label{fig:pred_pipe}   
\vspace{-5mm}
\end{figure}

\smartparagraph{Classification Accuracy.} \Cref{fig:pred_pipe} illustrates the \textit{classification} accuracy at the city and country levels and \textit{distance} threshold accuracy computation pipeline. For this evaluation, we benchmark on three datasets: GeoDE \cite{ramaswamy2023geode}, DollarStreet \cite{gaviria2022dollar}, and CityGuessr68k \cite{kulkarni2024cityguessr}. From GeoDE, we sampled 22k images based on 16 meta-tags having geolocalizable features. From DollarStreet, we manually sampled 1.3k images, removing indoor and non-geolocalizable samples. Since its metadata contains only country-level information, we evaluate this dataset solely for country classification. Additionally, we use the validation set of 14k images from CityGuessr and all 22k GeoDE samples for city and country classification tasks.

\vspace{-1mm}
\subsection{Results and Discussion}\label{sec:results}

\smartparagraph{GAEA-Bench Evaluation.}
\Cref{table:bench} presents the performance of 12 recent LMMs on GAEA-Bench. The results offer several insights: \myNum{i} Our proposed model, GAEA, achieves the highest average performance across decision-making questions (T/F and MCQs) and Short VQAs. Among proprietary models, GPT-4o \cite{gpt4} overall performs the best, with an accuracy of 60.3\%, excelling particularly in Long VQAs—outperforming GAEA by 1.9\% in this category. However, both open-source and proprietary models struggle with short-form questions. E.g., GPT-4o's accuracy drops from 66.3\% on long questions to 49.5\% on short questions. \myNum{ii} GAEA outperforms all LMMs with an average accuracy of 67.5\%, surpassing GPT-4o by 7.2\% and outperforming the second-best open-source model, LLaVA-OneVision \cite{llavaov}, by 18.2\%. \myNum{iii} Several open-source models, including LLaMA-3.2-11B \cite{llama3}, GLM-4V-9B \cite{glm}, and Phi-3.5-Vision \cite{abdin2024phi}, achieve comparable overall performance. \myNum{iv} LMMs perform better on decision-making questions (MCQs and T/F) than open-ended questions; see \Cref{fig:question_types}.
E.g., LLaVA-OneVision experiences a 26.5\% drop in accuracy on SVQA compared to MCQ questions. The low performance on free-form questions underscores the challenge of using short questions in effectively assessing conversational capabilities in the GAEA-Bench. \Cref{fig:qual_svqa,fig:qual_mcq,fig:qual_tf} shows comparisons with several LMMs.


\begin{table}[t]
\small
\centering
\footnotesize
\centering
\resizebox{0.49\textwidth}{!}{
\begin{tabular}{l ccccc}
\toprule
\textbf{Model Name} & \multicolumn{5}{c}{\textbf{Performance on Different QA Formats}} \\
\cmidrule(rl){2-6} 
 & \textbf{LVQA} & \textbf{SVQA}  & \textbf{MCQ} & \textbf{TF}  & \textbf{Average} \\

\midrule                     
    GeoChat-7B \cite{geochat}  & 23.9\dec{40.5}  &16.4\dec{36.2}  & 54.5\dec{18.2}  & 32.1\dec{46.4}  &  33.2\dec{34.3} \\
    LLaVA-Next-Mistral-7B \cite{liu2024llavanext}   & 48.0\dec{16.4}  & 23.2\dec{29.4}  & 29.2\dec{43.5} & 56.7\dec{21.8} & 37.7\dec{29.8}  \\
    Phi-3.5-Vision-Instruct \cite{abdin2024phi}     & 48.7\dec{15.7}  & 14.3\dec{38.3}  & 54.7\dec{18.0} & 57.4\dec{21.1} & 42.9\dec{24.6} \\
        LLaMA-3.2-Vision-11B \cite{llama32vision}   & 48.6\dec{15.8}  & 29.9\dec{22.7}   & 53.2\dec{19.5} & 47.1\dec{31.4} & 44.0\dec{23.5} \\
    GLM-4V-9B \cite{glm}           & 41.6\dec{22.8}  & 29.7\dec{22.9}  & 56.7\dec{16.0}  & 50.6\dec{27.9} & 45.2\dec{22.3} \\
    Qwen2.5-VL \cite{Qwen2.5-VL}  & 57.0\dec{7.4}  & 29.9\dec{22.7}  & 48.1\dec{24.6}  & 59.4\dec{19.1}  & 47.1\dec{20.4}  \\
    InternVL2-8B \cite{mplugowl3}     & 54.5\dec{9.9}  & 31.0\dec{21.6}  & 55.7\dec{17.0} & 56.8\dec{21.7} & 48.6\dec{18.9} \\
    LLaVA-OV-7B \cite{llavaov}    & 54.1\dec{10.3}  & 31.5\dec{21.1} & 58.0\dec{14.7}  & 56.4\dec{22.1} & 49.3\dec{18.2} \\
\midrule

Gemini-2.0-Flash \cite{team2023gemini} & 58.3\dec{6.1}   & 34.7\dec{17.9}  & 57.2\dec{15.5}  & 56.7\dec{21.8}  & 50.5\dec{17.0}  \\ 
GPT-4o-mini \cite{gpt4}~$\textsuperscript{*}$ & 61.8\dec{2.6}  & 34.1\dec{18.5}  & 54.9\dec{17.8}  & 33.9\dec{44.6}  & 43.4\dec{24.1} \\
GPT-4o \cite{gpt4}~$\textsuperscript{*}$ & \textbf{66.3}\textcolor{Green}{$_{+1.9}$}  & 49.5\dec{3.1}  & 59.4\dec{13.3}  & 69.6\dec{8.9}  & 60.3\dec{7.2} \\
\midrule
\rowcolor{green!20!} \textbf{GAEA~\textcolor{black}{(Ours)}}     & 64.4  &  \textbf{52.6} & \textbf{72.7} & \textbf{78.5} & \textbf{67.5}\\
\midrule
\end{tabular}
}
\vspace{-1mm}
\caption{\small{We benchmark 11 open-source and proprietary LMMs on GAEA-Bench. Notably, GAEA outperforms all open-source models and fares higher than the proprietary models on decision-making questions \textit{(MCQs and TFs)}. We provide the relative performance change for each model compared to GAEA. $\textsuperscript{*}$- We use GPT-4o as a judge for evaluation, and it has been documented that LLMs as judges prefer their long-form output~\cite{wataoka2025selfpreference, ye2025justice}, hence the scores for these models are likely overestimated.}}
\label{table:bench}
\vspace{-4mm}
\end{table}

\begin{figure}[t]
  \centering
  \includegraphics[width = \columnwidth]{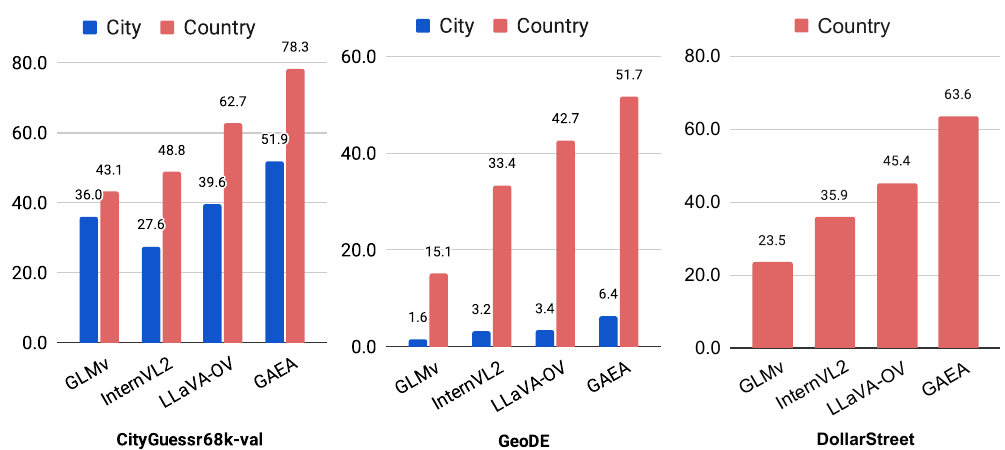}
  \vspace{-6mm}
   \caption{Classification accuracy for both city and country labels, where GAEA  surpasses several recent LMMs in performance.}
\label{fig:city_country_results}
\vspace{-6mm}
\end{figure} 

\smartparagraph{Standard Geo-localization Evaluation.}\label{sec:res_sota} \Cref{tab:benchmark_results} compares GAEA’s performance with various specialized encoder-only methods across three standard geolocalization benchmarks. While GAEA is trained on a large-scale conversational dataset with geolocalization capabilities, it achieves competitive results against specialized models. We evaluate against GaGA \cite{dou2024gaga}, which is trained on a dataset five times larger than ours, on IM2GPS and IM2GPS3k. However, we exclude comparisons on GWS15k due to differences in dataset curation. We contacted the authors of \cite{clark2023we} for the original GWS15k benchmark for fair evaluation.

\begin{table}[t]
\scriptsize
    \centering
    \scalebox{0.95}{
    \begin{tabular}{lp{1.8cm}p{0.7cm}p{0.8cm}p{0.8cm}p{0.9cm}}
        \toprule
        \textbf{Benchmark} & \centering \textbf{Model} & \centering \textbf{City} & \centering \textbf{Region} & \centering \textbf{Country} & \textbf{Continent} \\
        & & \textbf{25 km} & \textbf{200 km} & \textbf{750 km} & \textbf{2500 km} \\
        \midrule
        \multirow{10}{*}{IM2GPS \cite{hays2008im2gps}} 
        & PlaNet \cite{weyand2016planet} & 24.5 & 37.6 & 53.6 & 71.3 \\
        & CPlaNet \cite{cplanet} & 37.1 & 46.4 & 62.0 & 78.5 \\
        & ISNs \cite{isns} & 43.0 & 51.9 & 66.7 & 80.2 \\
        & TransLocator \cite{translocator} & 48.1 & 64.6 & 75.6 & 86.7 \\
        & GeoCLIP \cite{vivanco2024geoclip} & 41.8 & 60.8 & 77.2 & 89.9 \\
        & GeoDecoder \cite{clark2023we} & \textbf{50.2} & \textbf{69.0} & 80.0 & 89.1 \\
        & PIGEON \cite{haas2024pigeon} & 40.9 & 63.3 & \textbf{82.3} & \textbf{91.1} \\
        
        \arrayrulecolor{black}
        \cdashline{2-6}[2pt/2.5pt]
        & GeoReasoner \cite{georeasoner-icml25} & 24.9 & 48.1 & 65.8 & 82.3 \\
        & GaGA \cite{dou2024gaga} & 38.8 & 54.8 & 75.1 & 87.7 \\
        & \cellcolor{green!30}GAEA (Ours) & \cellcolor{green!30}43.0 & \cellcolor{green!30}57.4 & \cellcolor{green!30}77.2& \cellcolor{green!30}89.5 \\

        \midrule
        
        \multirow{10}{*}{IM2GPS3k \cite{vo2017revisiting}} 
        & PlaNet \cite{weyand2016planet} & 24.8 & 34.3 & 48.4 & 64.6 \\
        & CPlaNet \cite{cplanet} & 26.5 & 34.6 & 48.6 & 64.6 \\
        & ISNs \cite{isns} & 28.0 & 36.6 & 49.7 & 66.0 \\
        & TransLocator \cite{translocator} & 31.1 & 46.7 & 58.9 & 80.1 \\
        & GeoDecoder \cite{clark2023we} & 33.5 & 45.9 & 61.0 & 76.1 \\
        & GeoCLIP \cite{vivanco2024geoclip} & 34.5 & 50.7 & 69.7 & 83.8 \\
        & PIGEON \cite{haas2024pigeon} & 36.7 & 53.8 & 72.4 & 85.3 \\

        \arrayrulecolor{black}
        \cdashline{2-6}[2pt/2.5pt]
        & GeoReasoner \cite{georeasoner-icml25} & 26.5 & 40.4 & 57.7 & 72.8 \\
        & GaGA \cite{dou2024gaga} & 33.0 & 48.0 & 67.1 & 82.1 \\

    & \cellcolor{green!30}GAEA (Ours) & \cellcolor{green!30}\textbf{36.9} & \cellcolor{green!30}\textbf{56.0} & \cellcolor{green!30}\textbf{73.2} & \cellcolor{green!30}\textbf{86.7} \\

        \midrule
        

        

        

        

        
        \multirow{5}{*}{GWS15k \cite{clark2023we}} 
        & ISNs \cite{isns} & 0.6 & 4.2 & 15.5 & 38.5 \\
        & TransLocator \cite{translocator} & 1.1 & 8.0 & 25.5 & 48.3 \\
        & GeoDecoder \cite{clark2023we} & 1.5 & 8.7 & 26.9 & 50.5 \\
        & GeoCLIP \cite{vivanco2024geoclip} & 3.1 & {\textbf{16.9}} & {\textbf{45.7}} & {\textbf{74.1}} \\

        \arrayrulecolor{black}
        \cdashline{2-6}[2pt/2.5pt]
        & \cellcolor{green!30}GAEA (Ours) & \cellcolor{green!30}{\textbf{3.7}} & \cellcolor{green!30}16.7 & \cellcolor{green!30}43.3 & \cellcolor{green!30}73.5 \\
        \bottomrule
    \end{tabular}}
    \caption{\small{We benchmark the performance of various specialized models on standard geolocation datasets. GAEA demonstrates competitive results, outperforming GaGA on multiple distance thresholds in both IM2GPS and IM2GPS3k.}}
    \vspace{-5mm}
    \label{tab:benchmark_results}
\end{table}


On the IM2GPS3k benchmark, GAEA consistently outperforms all existing LMMs, including the domain-specialized PIGEON model \cite{haas2024pigeon}, across all four distance thresholds. Notably, GAEA surpasses PIGEON by 2.2\% at the 200 km threshold and by approximately 1.4\% at the continent level. Compared to GeoCLIP \cite{vivanco2024geoclip}, GAEA achieves gains of 5.3\% at the region level and 2.5\% at the city level. Additionally, GAEA significantly outperforms the open-source models, GaGA \cite{dou2024gaga}, with improvements of 6.1\% at the country level and 8.0\% at the regional level and GeoReasoner \cite{georeasoner-icml25} with 10.5\% at city, 15.6\% at region, and 13.9\% at a 2500 km threshold. On IM2GPS, GAEA outperforms both the specialized LMMs, GeoReasoner and GaGA on all four distance metrics. It also surpasses PIGEON and GeoCLIP models at the city level by ~2.1\% and ~1.2\%, respectively, while maintaining competitive performance across other thresholds. We also evaluate GAEA on GWS15k, one of the most challenging datasets, which includes non-geolocalizable landmarks. GAEA outperforms GeoCLIP \cite{vivanco2024geoclip} and GeoDecoder \cite{clark2023we} on city-level distance and achieves comparable performance at the region and country levels. \Cref{fig:city_country_results} presents GAEA's \textbf{Classification Accuracy} on three new datasets: CityGuessr68k-val \cite{kulkarni2024cityguessr}, GeoDE \cite{ramaswamy2023geode}, and DollarStreet \cite{gaviria2022dollar}. GAEA outperforms recent LMMs, including LLaVA-OneVision \cite{llavaov}, InternVL \cite{chen2024internvl}, and GLM-4V-9B \cite{glm}, on both city- and country-level classification. These results highlight GAEA’s extensive geographical coverage and strong geolocation capabilities.
\vspace{-0.7em}
\section{Conclusion}\label{sec:conclusion}

We introduced GAEA, the first interactive conversational model with specialized geolocation capabilities, explicitly trained on a large-scale conversational dataset, \datasetname. We meticulously designed the dataset to enhance GAEA’s reasoning, conversational abilities, and geolocation accuracy. We curated geolocalizable images from MP-16, GLDv2, and CityGuessr68k, enriching them with auxiliary context and metadata, such as geographic clues and climate zones. In addition to a high-quality instruction set, we present GAEA-Bench, a comprehensive benchmark that evaluates LMMs across multiple question types, including MCQs, True/False, short- and long-VQAs. Our results show that GAEA outperforms recent LMMs on GAEA-Bench, demonstrating strong geolocation and conversational capabilities by leveraging OpenStreetMap (OSM) data. These findings establish GAEA  as a strong baseline for future research in geolocalization.

\section{Acknowledgement}
\noindent This work was supported by ``MFC Lockheed Martin, Orlando". We would also like to thank David Shatwell, Manu S Pillai, Praveen Tirupattur, Brian Dina, and Suranadi Dodampaganmage for their insightful discussions and contributions. 
{
    \small
    \bibliographystyle{ieeenat_fullname}
    \bibliography{main}
}
\clearpage
\maketitlesupplementary



\noindent We organize the  Supplementary Material as follows: In \Cref{sec:rationale}, we provide additional details of our dataset, \datasetname\ . In \Cref{sec:baseline_appendix}, we summarize baselines we compared against, describe prompts used during training, inference and evaluation, and provide training details and additional results on \textnormal{\modelname}. 

\begin{figure}[!h]
  \centering
  \includegraphics[width=0.4\textwidth]{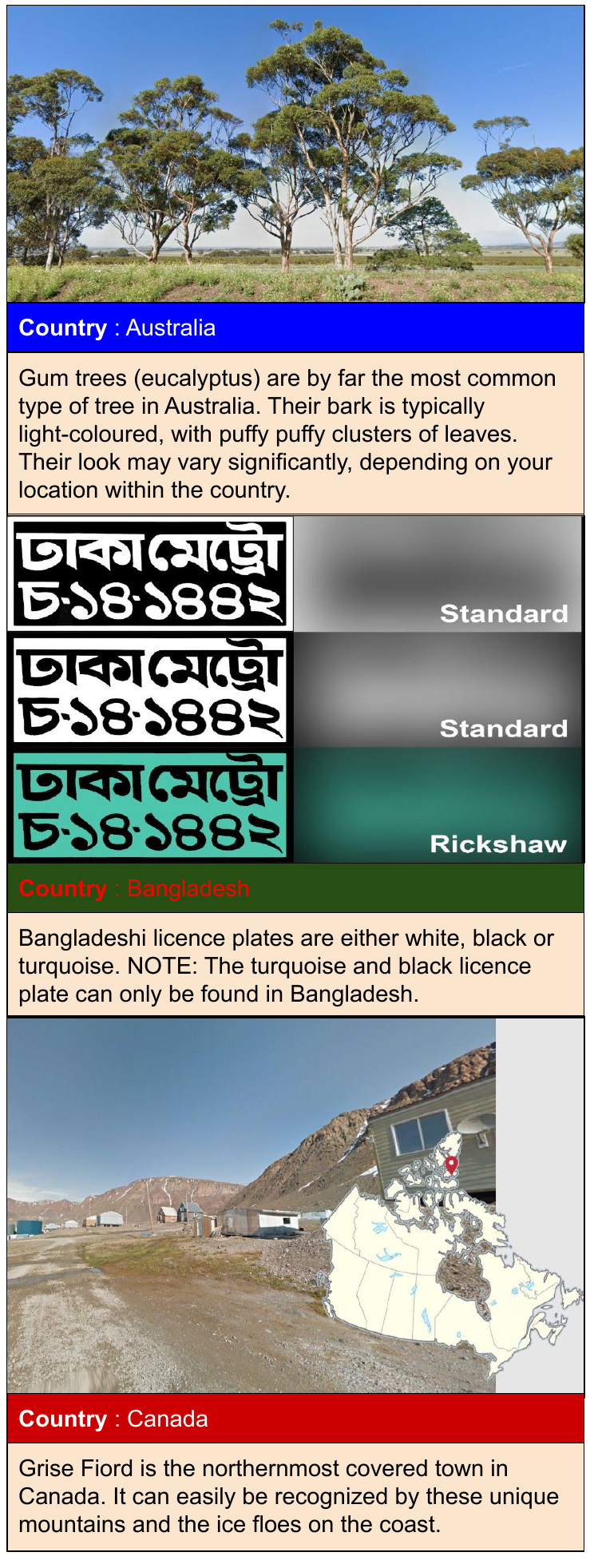}
  \caption{Examples of country-specific geographical clues used in our dataset. These clues are used to support geographically grounded reasoning in our QA generation.}
  \label{fig:clues}
\end{figure}

\section{Addendum to the Dataset}\label{sec:rationale}
In this section, we present the dataset statistics and challenges encountered in its creation. Additionally, we discuss our plans to address these limitations in future works.

\begin{figure*}[t]
  \centering
  \includegraphics[width=\textwidth]{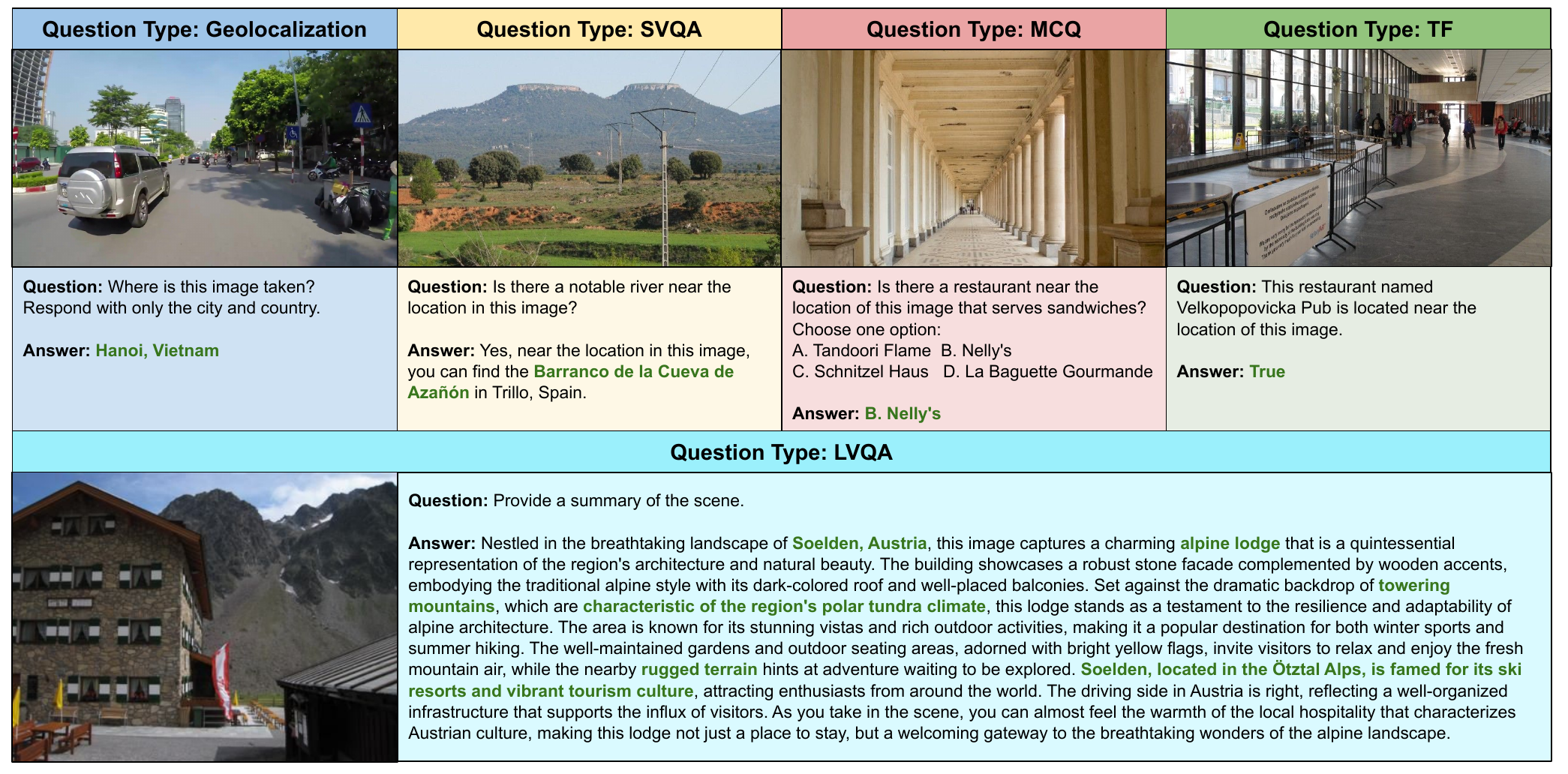}
  \caption{Examples of the four question types in our dataset: SVQA, MCQ, TF, and LVQA. Each type targets a distinct reasoning skill grounded in geographical, visual, or contextual understanding. Our dataset has three categories, including Geolocalization, Reasoning(LVQA), and Conversational (SVQA, MCQ, TF) QAs, as shown in the figure.}
  \label{fig:training_examples1}
\end{figure*}



\subsection{GeoGuessr Context Clues}
To support geographically-grounded reasoning in our dataset, we incorporate country-specific context clues inspired by the popular game GeoGuessr \cite{geoguessr}. We web-crawled and extracted 129 high-quality clues from Plonkit \cite{plonkit}, a community-driven open-source resource widely used by over 65 million GeoGuessr players. To improve country coverage and address missing entries (e.g., France, New Zealand), we augmented the set with 58 additional GPT-4o-generated clues, carefully aligned with Plonkit’s descriptive style. This expanded our coverage to 187 countries. These clues highlight distinctive, often visual, geographical features, such as Bangladesh's unique turquoise license plates, Australia’s widespread gum trees, or the ice-covered terrain of Grise Fiord in Canada (see Figure~\ref{fig:clues}). These examples are leveraged to generate reasoning-based multiple-choice and open-ended questions. Furthermore, we enrich each sample with auxiliary metadata: Köppen-Geiger climate classification \cite{beck2018present}, traffic orientation from WorldStandards \cite{ws}, land cover data from EarthEnv \cite{earthenv}, and scene labels derived from the Places2 database \cite{zhou2017places}.

\subsection{Challenges with Open Street Maps (OSM)} \label{sec:OSM}
OpenStreetMaps (OSM) \cite{osm} is a rich data source for geospatial applications. It contains a wide variety of geographic and infrastructure-related information. Using such a vast open-source dataset, we can collect data about stationary objects in the world, including infrastructure, topological information, various types of amenities (e.g., schools, hospitals, restaurants), transportation networks, international country boundaries, historical and cultural sides, and natural features (e.g., forests, rivers, and seas). Each feature from the OSM dataset has several associated features, such as names and physical characteristics. In \datasetname, we geocode the visual sample with its GPS coordinates and use the location information (longitude and latitude) as a query to the OSM database to fetch geospatial information within a ~1 KM radius and further utilize that information to generate question-answer pairs for the training of \textnormal{\modelname}.

Despite being such a rich source of data, OSM faces several challenges. One major issue is the variability in data quality and completeness, as contributions to OSM are made by the open-source community, which may result in inconsistent information across different regions. Urban areas often have much more detailed information than rural areas, leading to less comprehensive annotations for rural regions. Another inconsistency related to human annotations stems from the different representations of the same label in different areas, introducing inherent heterogeneity in the structure of OSM data. For instance, some users might label a path as a ``trail," while others might call it a ``footway," and distinctions between what counts as a ``park" versus a "garden" are not always clear. Moreover, querying and retrieving data from OSM is a compute-intensive task. It often becomes slower as the number of queries increases and struggles to handle dense or redundant information, necessitating efficient filtering and optimization techniques. Lastly, the information is not always up-to-date, as volunteers update different areas at different times. While some locations may have very recent data, others may be outdated, and sometimes different parts of the same area may contain information from varying periods.


\subsection{Training Examples and Data Statistics}\label{sec:stat}
\datasetname\ covers 234 different countries and territories, and 41,481 cities. 
\Cref{tab:stats} shows the statistics of GAEA-1.4M in detail. \Cref{fig:training_examples1} shows some qualitative examples of various question types in our \datasetname\, leveraging OSM, and GeoGuessr geographical clues for constructing conversational QA pairs.

\begin{table}[h]
\scriptsize
    \centering
    \resizebox{0.47\textwidth}{!}{
    \begin{tabular}{lp{1.7cm}p{0.7cm}p{0.8cm}p{0.8cm}p{0.9cm}}        \toprule
    Total Images & 822,951 \\ 
    Total Cities / Countries & 41,481 / 234 \\
    Total Questions & 1,432,519 \\ 
    Total Geo-Loc QAs & 822,951 \\
    Total LVQAs & 236,935 \\
    Total SVQAs & 267,668 \\
    Total MCQs & 48,673 \\ 
    Total True/False QAs & 56,292 \\
    \hline
    \end{tabular}}
    \caption{Dataset Statistics}
    \label{tab:stats}
    \vspace{-4.2mm}
\end{table}

\begin{figure*}
  \centering
  \includegraphics[width=\linewidth]{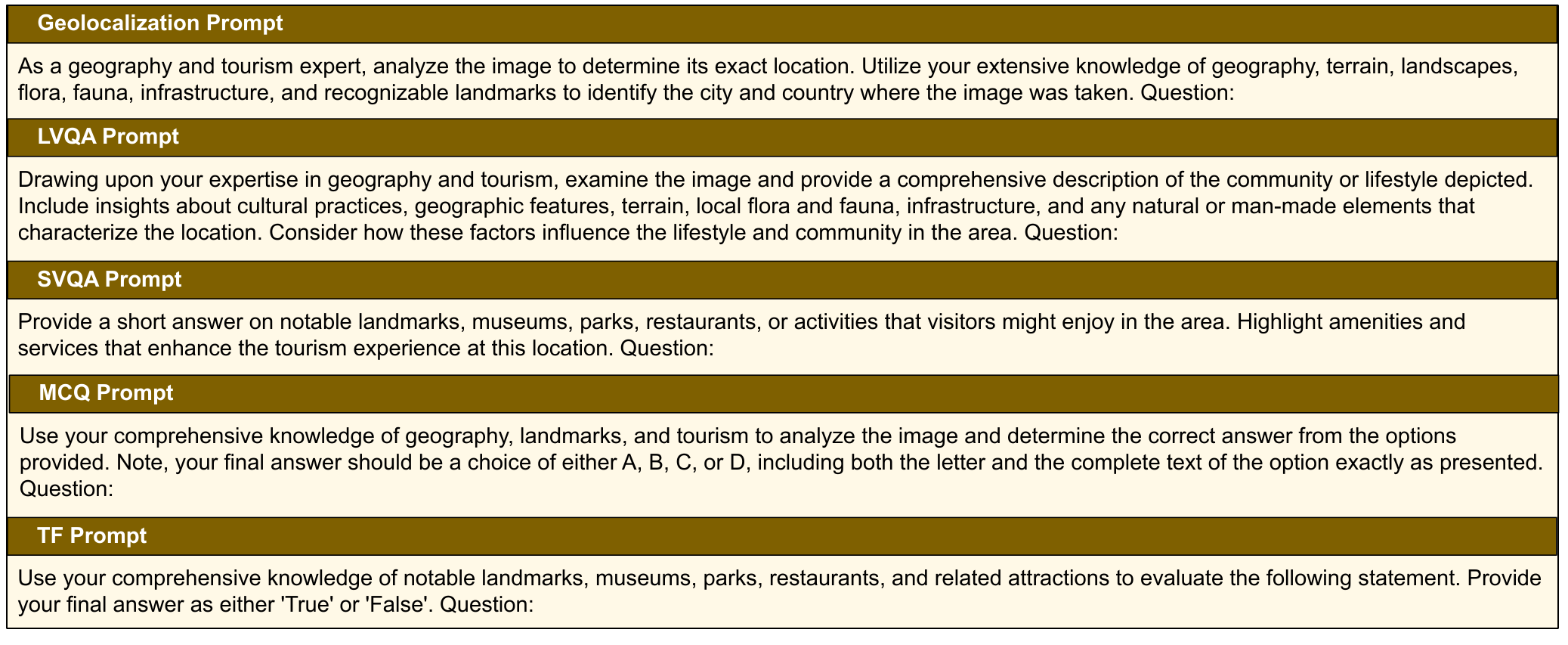}
  \caption{Task-specific training prompts used to instruct the model for each of the five question types in GAEA. Each prompt is carefully designed to elicit a targeted form of geographical reasoning. These prompts ensure consistent and interpretable outputs during both the training and evaluation phases while training GAEA.}
  \label{fig:train_prompts}
\end{figure*}



\section{Addendum to Baseline and Evaluation}\label{sec:baseline_appendix}
This section covers the models used for comparison with \modelname, the prompts used during training and inference, the prompts used for evaluating \dataseteval, and the training hyperparameters.

\subsection{Baselines}\label{sec:baselines}
We benchmark 8 top-performing open-source LMMs, including LLaMA 3.2-Vision \cite{llama32vision}, InternVL2 \cite{chen2024internvl}, Qwen2.5-VL \cite{Qwen2.5-VL}, Phi3.5-vision-instruct \cite{abdin2024phi}, GeoChat \cite{geochat}, LlaVA-OneVision \cite{llavaov}, GLM-4V-9B \cite{glm}, LLaVA-NeXT-Mistral-7B \cite{liu2024llavanext}, and 3 proprietary models, Open-AI's GPT-4o, GPT-4o-mini \cite{gpt4}, and Google's Gemini-2.0-Flash \cite{team2023gemini} on \dataseteval. 

Additionally, we compared the performance of \textnormal{\modelname} against six state-of-the-art (SoTA) specialized geolocalization models, namely PlaNet \cite{weyand2016planet}, CPlaNet \cite{cplanet}, ISNs \cite{isns}, TransLocator \cite{translocator}, GeoDecoder \cite{clark2023we}, and PIGEON \cite{haas2024pigeon} and open-source geolocalization LMMs GeoReasoner \cite{georeasoner-icml25} and GaGA \cite{dou2024gaga} on three standard geo-localization benchmarks including IM2GPS \cite{hays2008im2gps}, IM2GPS3k \cite{vo2017revisiting}, and GWS15k \cite{clark2023we}. We also compare our city-country classification performance with other LMMs on 3 benchmarks CityGuessr68k \cite{kulkarni2024cityguessr}, GeoDE \cite{ramaswamy2023geode} and DollarStreet \cite{gaviria2022dollar}. Preprocessing for these benchmarks is described in \Cref{sec:eval} in the main paper. 
\begin{figure*}
  \centering
  \includegraphics[width=\linewidth]{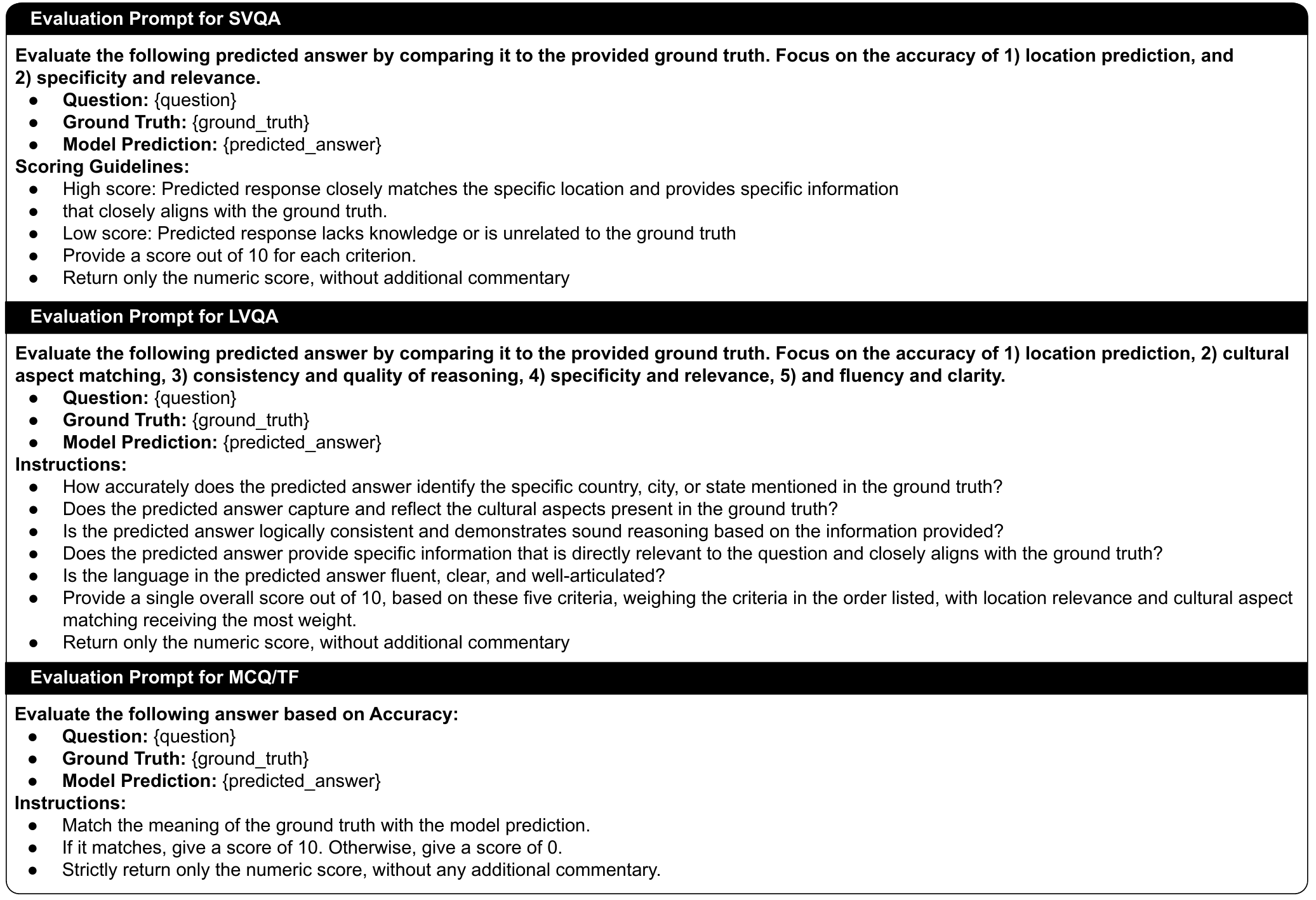}
  \caption{Evaluation prompts used for GPT-based scoring of model predictions across SVQA, LVQA, MCQ, and TF question types in the GAEA-Bench. Each prompt specifies detailed criteria, such as location accuracy, cultural relevance, specificity, and fluency, used to assign a numeric score for qualitative assessment.}
  \label{fig:prompts}
\end{figure*}

\subsection{Prompts Used During Training and Inference}\label{sec:appendix_prompt}
When training {\modelname}, we employed the task-specific prompts shown in \Cref{fig:train_prompts} to align the model's understanding with target objectives. During inference, these identical prompts were used for all models evaluated on \dataseteval\ to ensure comparability.

\subsection{Prompts Used in Evaluation}\label{sec:appendix_eval_prompt}
Figure~\ref{fig:prompts} presents the task-specific prompts used for evaluating model-generated answers via GPT-based assessment. For SVQA and LVQA, the prompts emphasize multi-faceted criteria such as location accuracy, specificity, cultural relevance, and reasoning quality, encouraging a nuanced evaluation of open-ended responses. In contrast, MCQ and TF tasks are scored based on strict binary accuracy by matching the predicted answer to the ground truth. These structured prompts ensure consistent, interpretable, and criterion-aligned evaluation across different QA types.

\begin{table}
\scriptsize
    \centering
    \scalebox{1.3}{
    \begin{tabular}{lp{1.7cm}p{0.7cm}p{0.8cm}p{0.8cm}p{0.9cm}}        \toprule
    Number of epochs                  & 1               \\
    Global batch size            & 128               \\
    Gradient accumulation steps & 4               \\
    Initial learning rate       & $10^{-5}$            \\
    Learning rate scheduler     & cosine          \\
    Warmup ratio                & 0.03            \\
    LoRA rank ($r$)                   & 16               \\
    LoRA alpha ($\alpha$)                  & 32               \\
    LoRA dropout                & 0.01            \\
    Precision             & \texttt{bfloat16}        \\
    \hline
    \end{tabular}}
    \caption{Hyperparameters used for training \modelname.}
    \label{table:hyper}
\end{table}

\subsection{Training Hyperparameters}\label{sec:hyper}
We perform single-stage training on the baseline, Qwen2.5-VL \cite{Qwen2.5-VL} using \datasetname. The training is conducted for 1 epoch with a global batch size of 128, accumulating gradients every 4 steps. The initial learning rate is set to $10^{-5}$, using a cosine learning rate scheduler to provide a smooth decay in the learning rate. The warmup ratio is configured at 0.03. We performed LoRA\cite{lora}-finetuning with a rank, $r=16$, $\alpha=32$, and a dropout rate of 0.01. The model operates in \texttt{bfloat16} precision. We also use flash attention \cite{dao2022flashattention}. We list the training hyperparameters in the Table \ref{table:hyper}. 
\begin{figure*}[t]
  \centering
  \includegraphics[width=\linewidth]{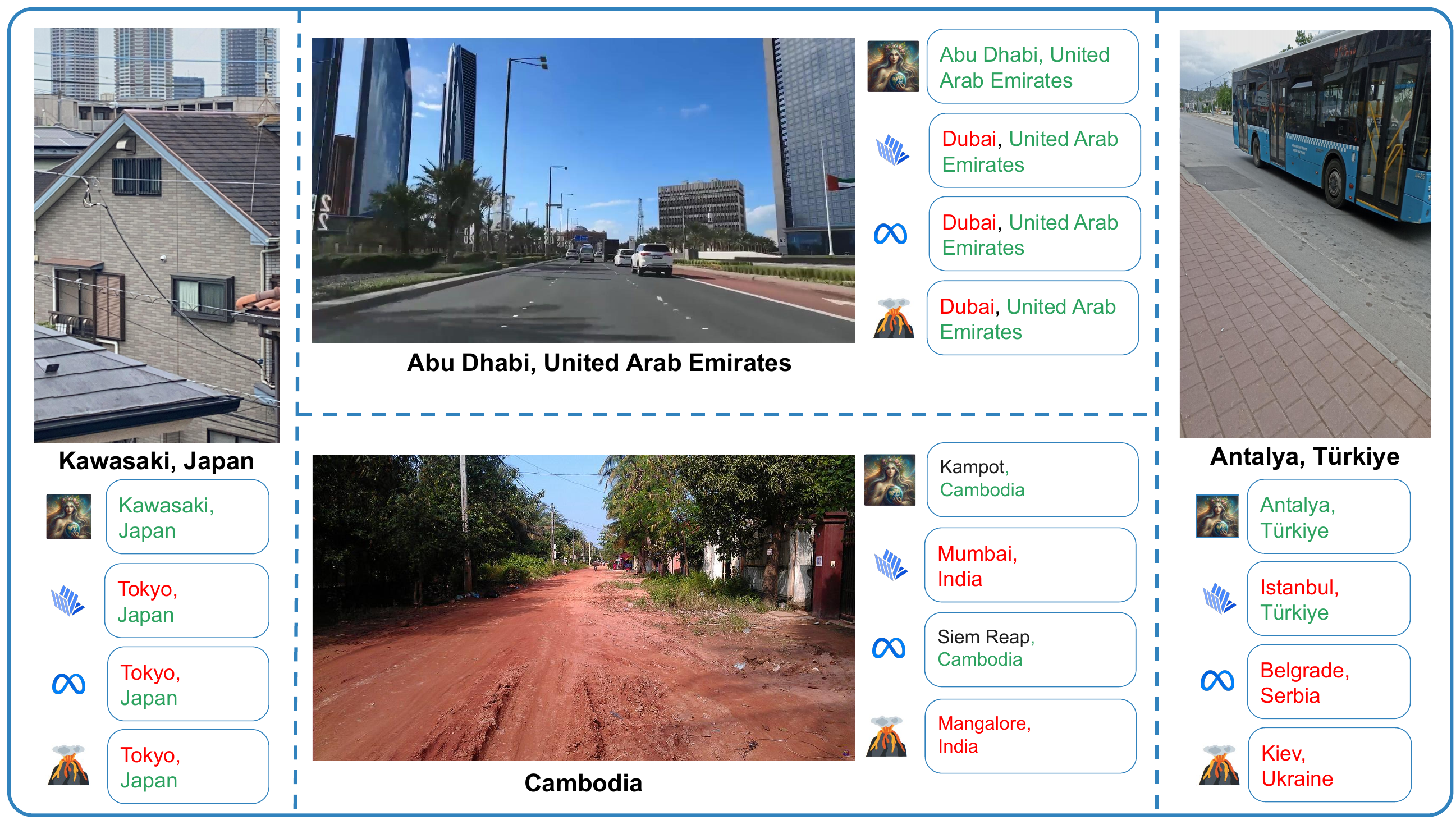}
  \caption{Qualitative Geolocalization comparison showing GAEA’s ability to predict accurate locations across varied types of scenes and landscapes around the world.}
  \label{fig:qual_cc}
\end{figure*}

\begin{figure*}
  \centering
  \includegraphics[width=\textwidth]{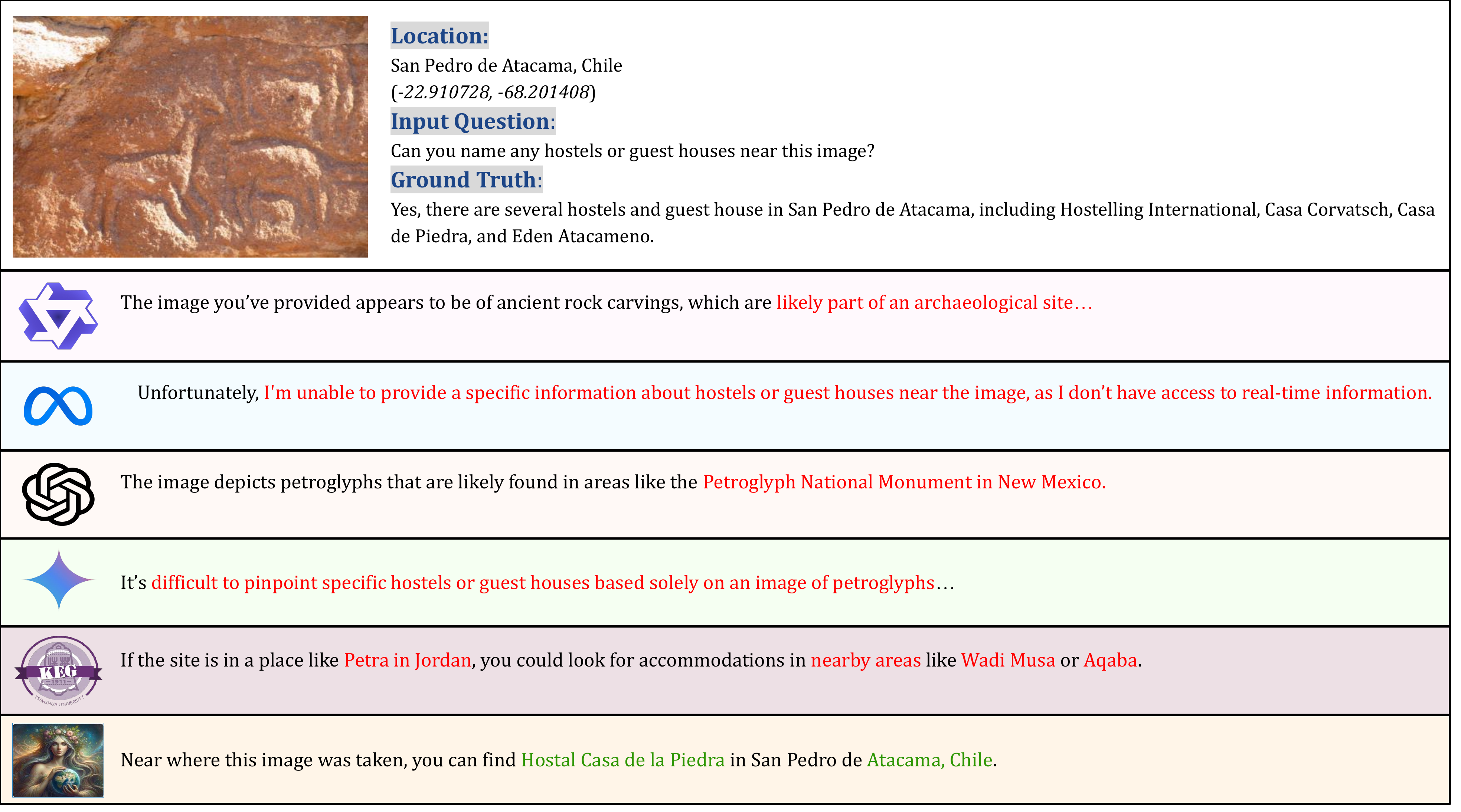}
  \caption{Qualitative SVQA comparison showing GAEA’s ability to provide accurate, location-specific answers where other LMMs fail.}
  \label{fig:qual_svqa}
\end{figure*}
\begin{figure*}
  \centering
  \scalebox{1}{\includegraphics[width=\linewidth]{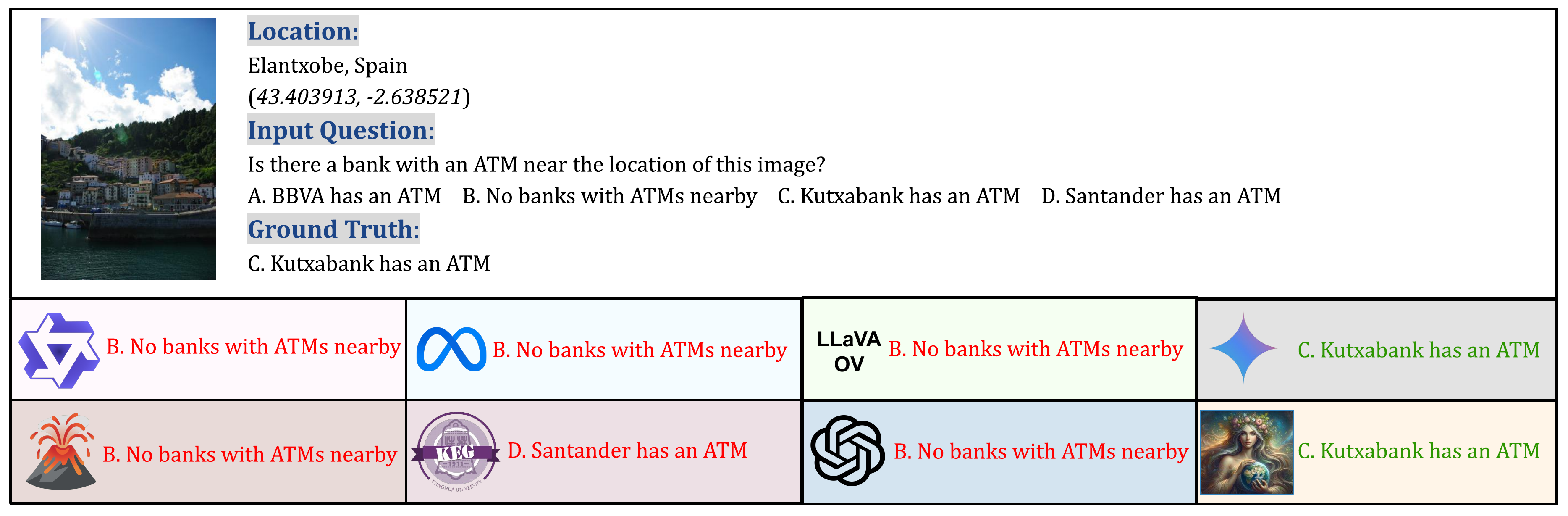}}
  \caption{Qualitative MCQs comparison showing GAEA’s ability to provide accurate answers where other LMMs fail.}
  \label{fig:qual_mcq}
\end{figure*}
\begin{figure*}
  \centering
  \scalebox{1}{\includegraphics[width=\linewidth]{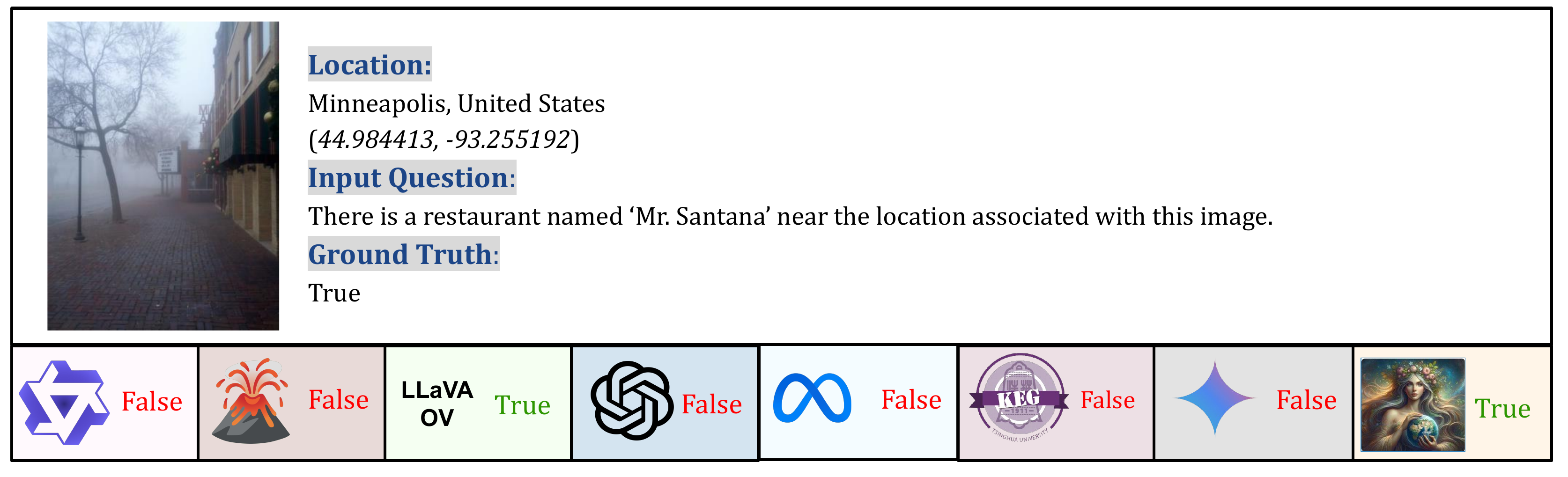}}
  \caption{Qualitative True/False comparison showing GAEA’s ability to provide accurate answers where other LMMs fail.}
  \label{fig:qual_tf}
\end{figure*}
\begin{figure*}[!t]
  \centering
  \scalebox{1}{\includegraphics[width=\linewidth, height=3in]{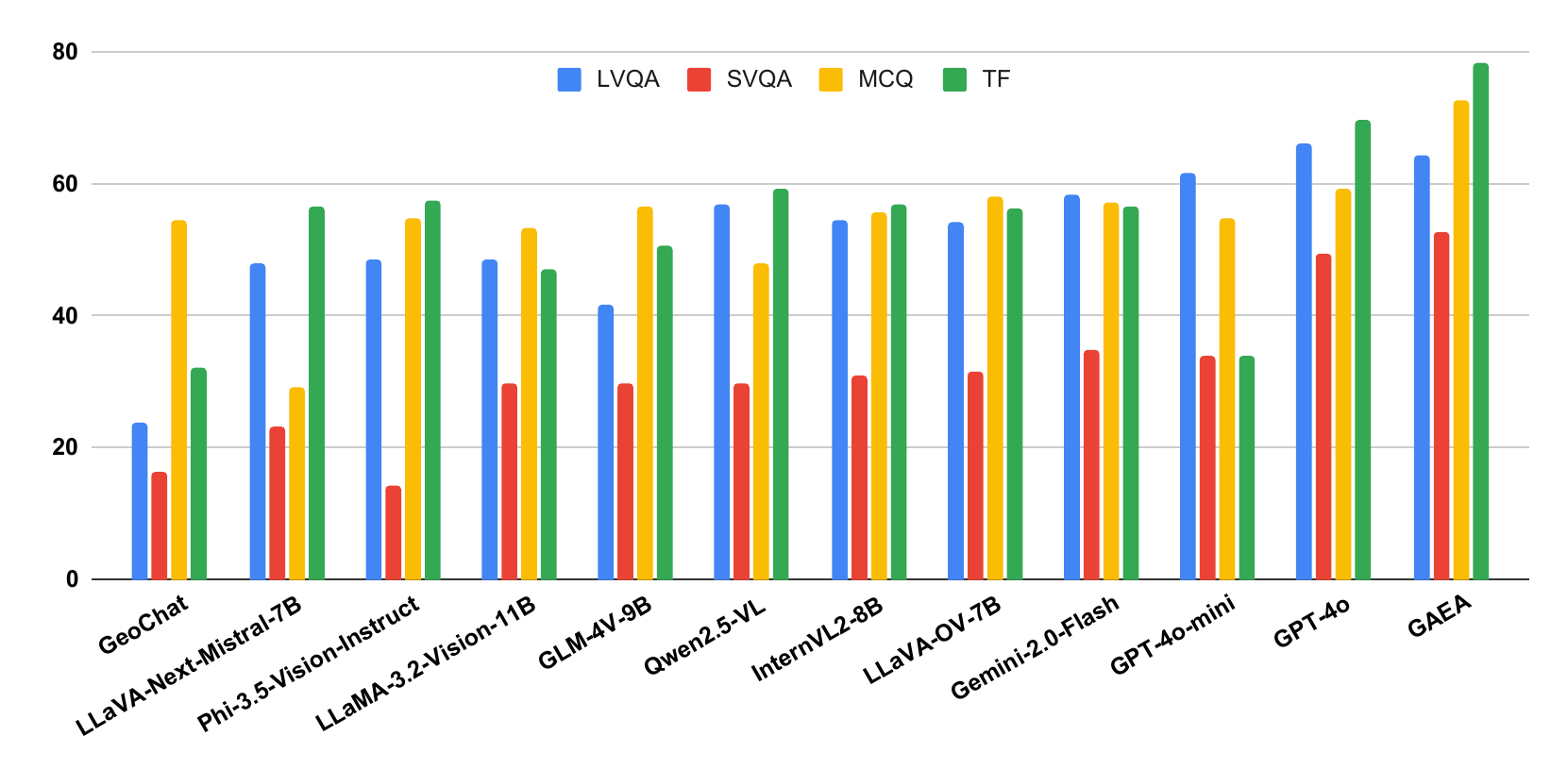}}
  \caption{We showcase the performance of various LMMs on four diverse question types. GAEA outperforms on average across all question forms.}
  \label{fig:question_types}
\end{figure*}

\subsection{Additional Results}\label{sec:app_qual}
In this Section, we discuss additional qualitative results of \textnormal{\modelname} and compare them with selected open-source and proprietary models (as mentioned in \Cref{table:bench} in the main paper). \Cref{fig:qual_cc} presents a comparison of city-country predictions against other competing models. We also show the qualitative results of \textnormal{\modelname} on short questions (SVQA), multiple-choice questions (MCQs), and true or false questions (TF) in \Cref{fig:qual_svqa,fig:qual_mcq,fig:qual_tf}. For these Figures, we highlight correct predictions with \textcolor{codegreen}{green}, while incorrect predictions are marked as \textcolor{red}{red}. Quantitative results on GAEA-Bench are summarized in \Cref{fig:question_types}.

\end{document}